\documentclass[10pt,twocolumn,letterpaper]{article}

\def\cvprPaperID{***}  
\def\confName{CVPR}
\def\confYear{2026}
\makeatletter
\providecommand{\paperID}{\cvprPaperID}
\makeatother

\PassOptionsToPackage{table}{xcolor}

\usepackage[final]{cvpr}
\usepackage[utf8]{inputenc}
\usepackage[T1]{fontenc}
\usepackage{url}
\usepackage{booktabs}
\usepackage{amsmath, amssymb, amsfonts}
\usepackage{graphicx}
\usepackage{subcaption}
\usepackage{placeins}
\usepackage{float}
\usepackage{xcolor}
\usepackage{comment}
\usepackage{import}
\usepackage{tikz}
\usepackage{enumitem} 
\usepackage{dashrule}
\usepackage{stfloats}
\usepackage{setspace}
\usepackage{amsmath}
\usepackage{amssymb}
\usepackage{booktabs}
\usepackage{multirow}

\definecolor{cvprblue}{rgb}{0.21,0.49,0.74}

\definecolor{lightyellow}{rgb}{1.0, 1.0, 0.8}
\definecolor{bestrow}{gray}{0.92}

\usepackage{bm}

\usepackage[linesnumbered,algoruled,noend,noline]{algorithm2e}
\SetInd{0.3em}{0.3em}

\usepackage{booktabs, array, tabularx}

\newcommand{\Ccal}{\mathcal{C}}

\newcommand{\Fcal}{\mathcal{F}}
\newcommand{\Gcal}{\mathcal{G}}

\newcommand{\Ical}{\mathcal{I}}

\newcommand{\Lcal}{\mathcal{L}}
\newcommand{\Mcal}{\mathcal{M}}

\newcommand{\Qcal}{\mathcal{Q}}

\defcitealias{Oquab:arXiv:2023:DINOv2}{DINOv2}
\defcitealias{Zhang:CVPR:2024:TellingLeftRight}{TellingLeftRight}
\defcitealias{Mariotti:CVPR:2024:ViewpointSphereMap}{SphereMap}

\subimport{./}{macros}

\usepackage[pagebackref,breaklinks,colorlinks]{hyperref}
\hypersetup{
    colorlinks=true, 
    linkcolor=blue,
   filecolor=magenta, 
    urlcolor=blue,    
      citecolor=blue 
}

\renewcommand{\sectionautorefname}{\S} 
\renewcommand{\subsectionautorefname}{\S}

\title{Cross-Instance Gaussian Splatting Registration via\\ Geometry-Aware Feature-Guided Alignment}

\author{
Roy Amoyal \quad Oren Freifeld \quad Chaim Baskin\\
Ben-Gurion University of the Negev, Israel
}

\begin{document}

\maketitle

\begin{figure*}[t]
  \centering
  \newcommand{\subh}{0.35\textwidth}
  \scalebox{0.9}[0.85]{
  \begin{minipage}[t]{0.47\textwidth}
    \centering
    \textbf{(A) Cross-instance registration}\par\vspace{4pt}
    \subcaptionbox{\footnotesize Random Initialization: view 1}[0.485\linewidth]{
\includegraphics[height=0.76\linewidth,width=0.9025\linewidth]{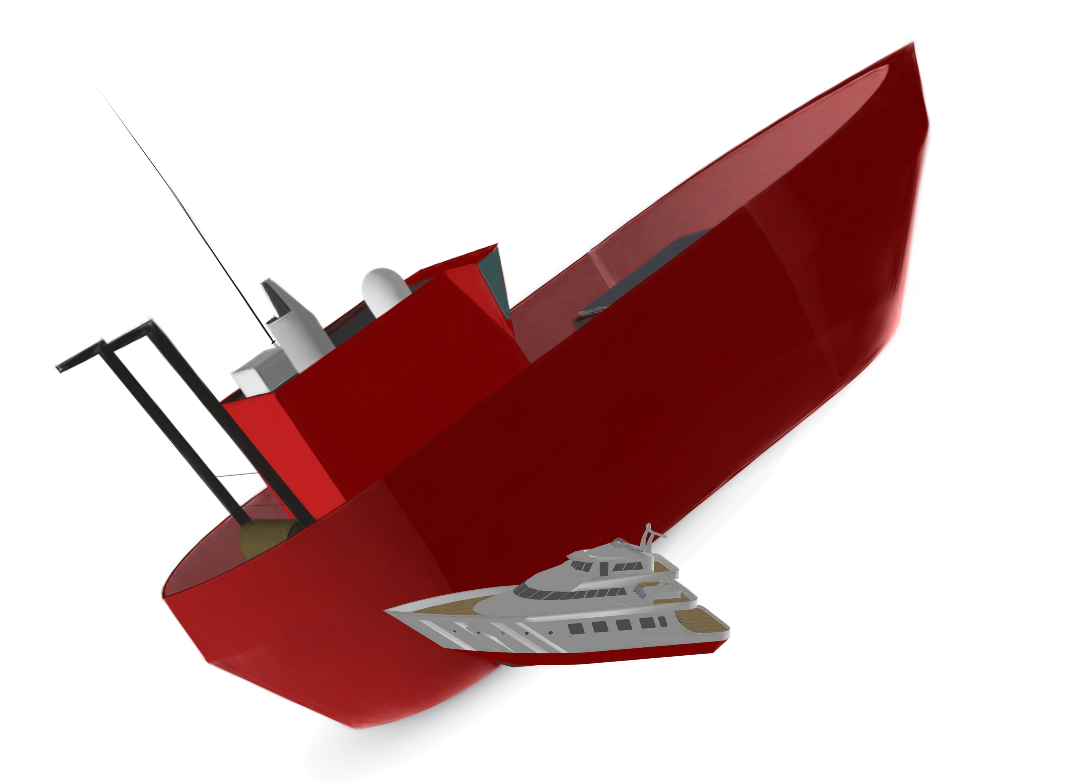}\vspace{-2mm}}
    \hfill
    \subcaptionbox{\footnotesize { GSA (ours) Result: view 1}}[0.485\linewidth]{
\includegraphics[height=0.76\linewidth,width=0.9025\linewidth]{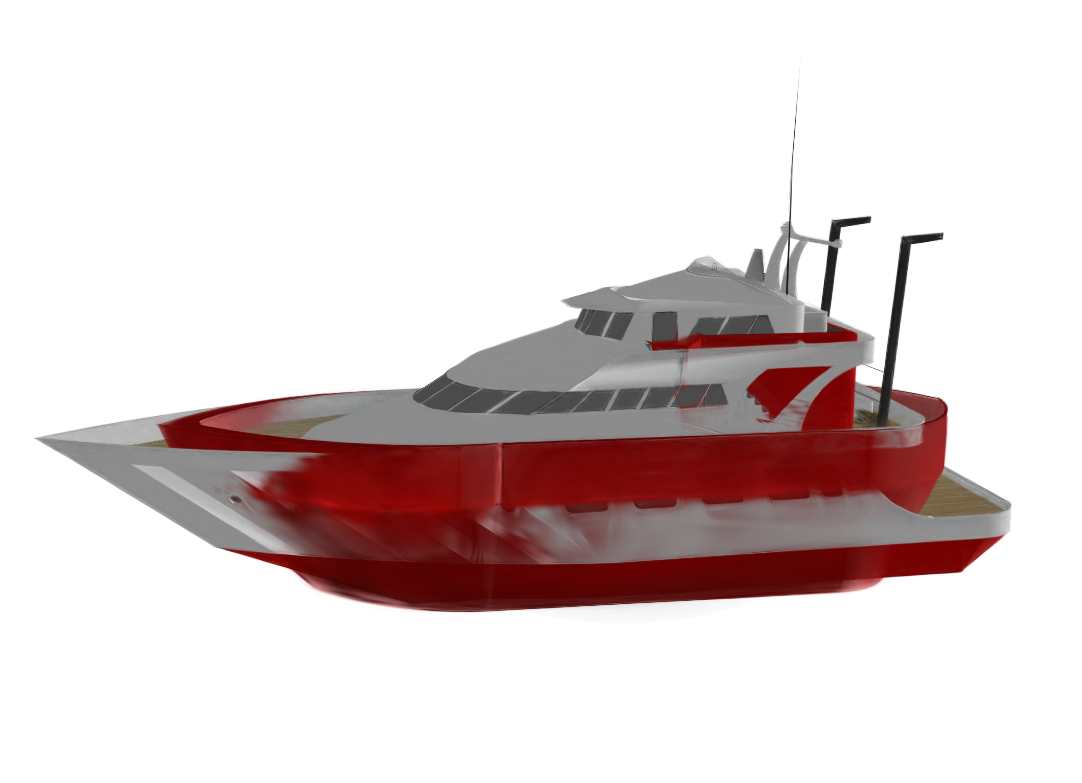}
      \vspace{-3mm}}
    \vspace{4pt}
    \subcaptionbox{\footnotesize Random Initialization: view 2}[0.485\linewidth]{
\includegraphics[height=0.76\linewidth,width=0.9025\linewidth]{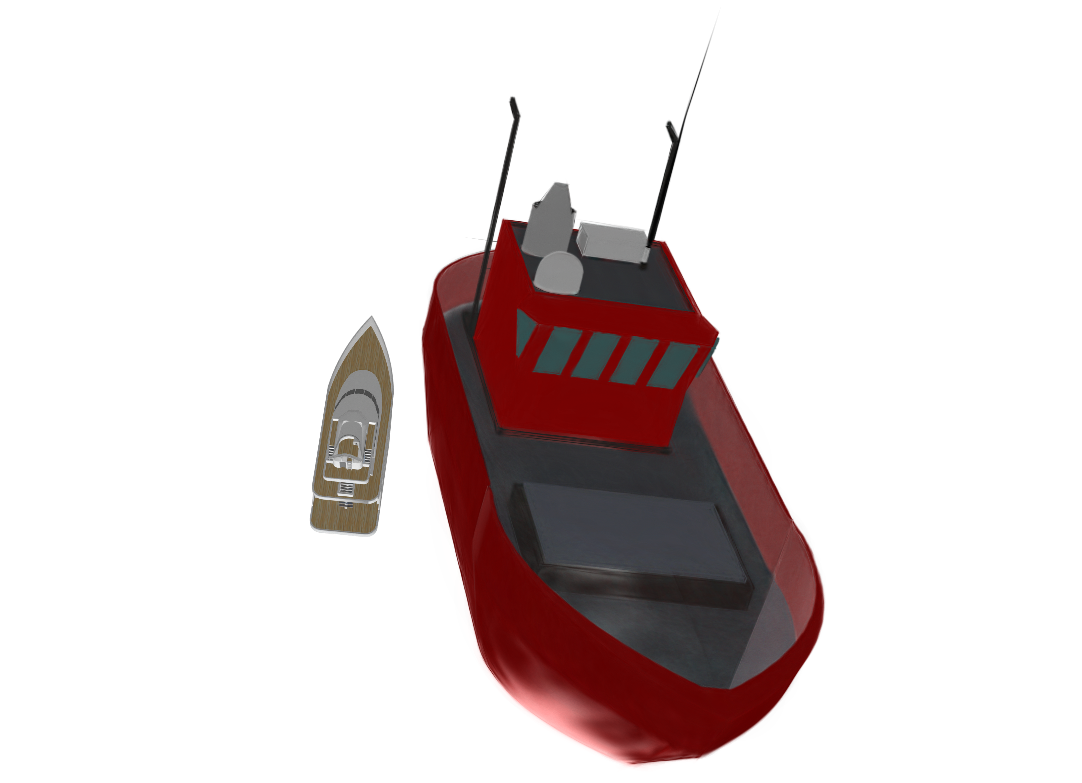}\vspace{-1mm}}
    \hfill
    \subcaptionbox{\footnotesize {GSA (ours) Result: view 2}}[0.485\linewidth]{
      \includegraphics[height=0.76\linewidth,width=0.9025\linewidth]{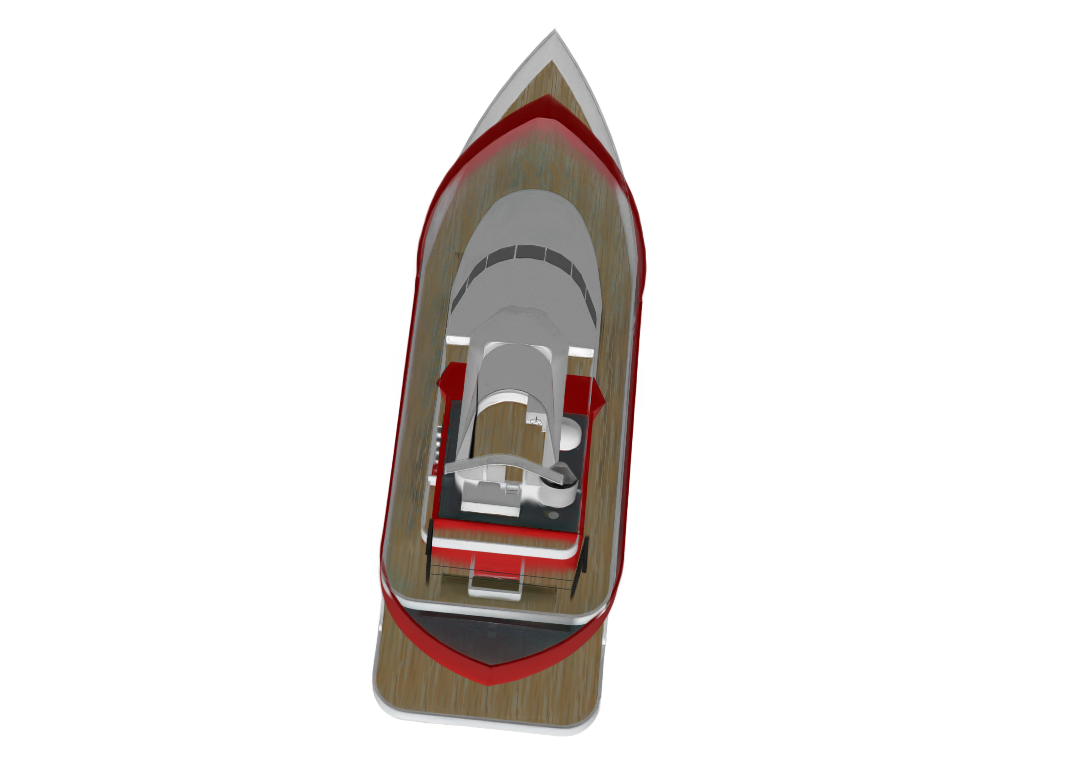}
      \vspace{-1mm}}
  \end{minipage}
  \hfill
  \begin{minipage}[t]{0.02\textwidth}
    \centering
    \begin{tikzpicture}[baseline=(current bounding box.north)]
      \draw[dash pattern=on 3pt off 3pt, gray!70, line width=0.6pt] (0,0) -- (0,-6.5cm);
    \end{tikzpicture}
  \end{minipage}
  \hfill
  \begin{minipage}[t]{0.47\textwidth}
    \centering
    \textbf{(B) Category geometric-consistent object replacement}\par\vspace{4pt}
    \subcaptionbox{\footnotesize Source Object}[0.485\linewidth]{
      \includegraphics[height=\subh,width=\linewidth,keepaspectratio]{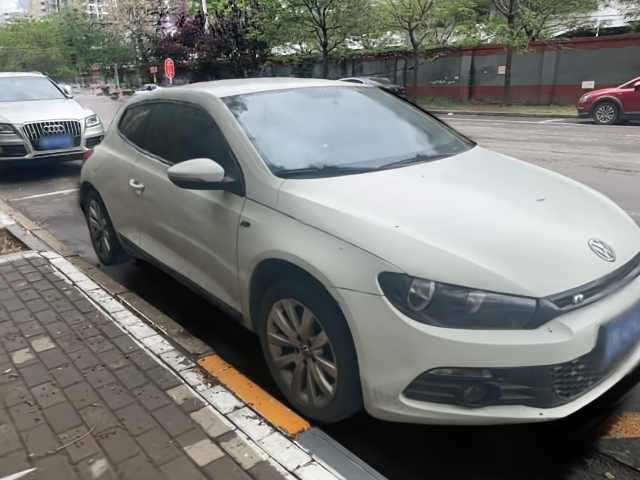}}
    \hfill
    \subcaptionbox{\footnotesize  Intra-dataset Replacement}[0.485\linewidth]{
      \includegraphics[height=\subh,width=\linewidth,keepaspectratio]{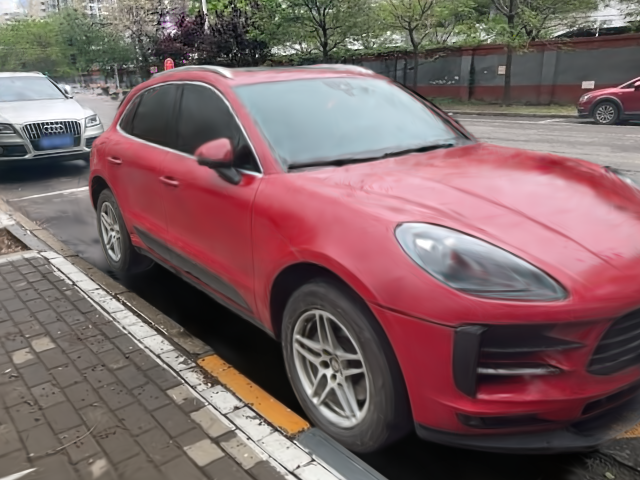}}
    \vspace{4pt}
    \subcaptionbox{\footnotesize Intra-dataset Replacement}[0.485\linewidth]{
      \includegraphics[height=\subh,width=\linewidth,keepaspectratio]{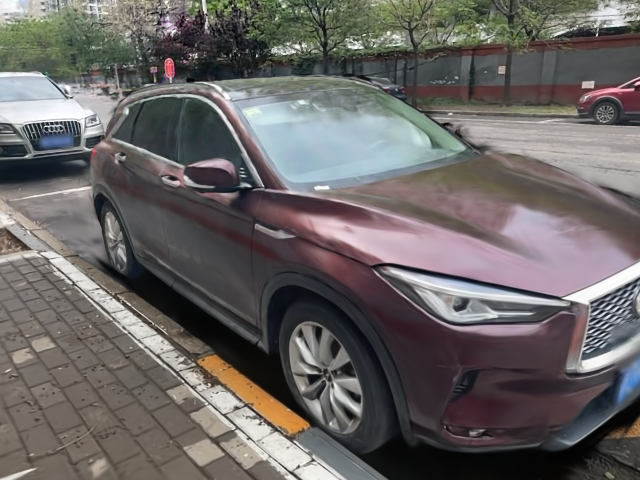}}
    \hfill
    \subcaptionbox{\footnotesize  Inter-dataset Replacement}[0.485\linewidth]{
      \includegraphics[height=\subh,width=\linewidth,keepaspectratio]{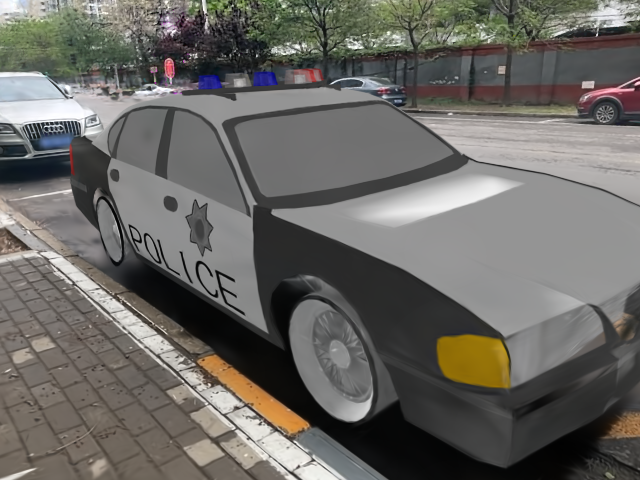}}
  \end{minipage}
} \vspace{-1mm}
  \caption{
(A) Cross-instance registration: GSA successfully aligns the red boat to the white boat despite extremely poor initialization, including a 180° rotation and over 5× scale difference.
(B) GSA enables geometrically consistent object replacement even across datasets, e.g., replacing the white car from \emph{3D Real Car}~\cite{du:Arxiv:2024:3DRealCar} with purple/red cars from the same dataset and a police car from \emph{ShapeNet}~\cite{chang:Arxiv:2015:Shapenet}.
}
  \label{fig:intro_split}
\end{figure*}
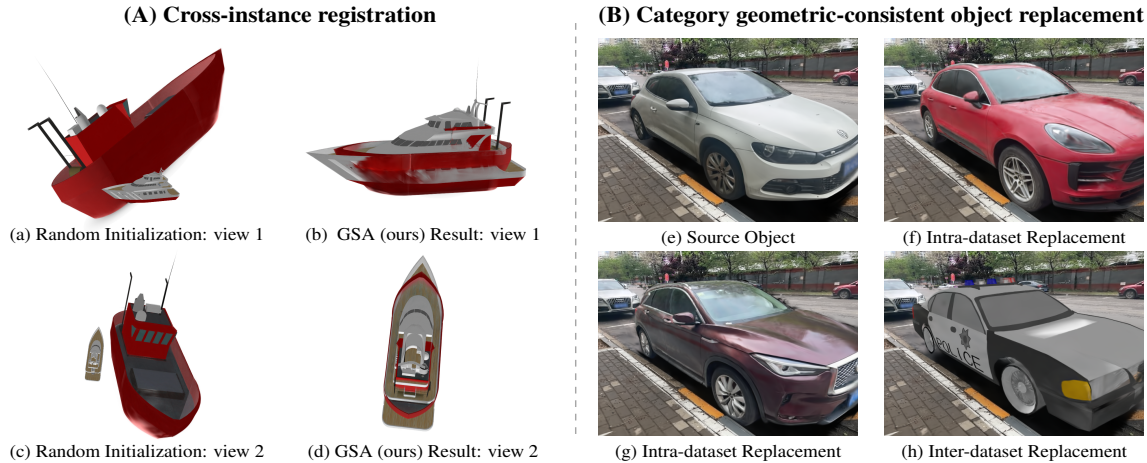

\begin{abstract}
We present \emph{Gaussian Splatting Alignment} (GSA), a novel method for aligning two  independent 3D Gaussian Splatting (3DGS) models via a similarity transformation (rotation, translation, and scale), even when they are of \emph{different} objects in the same category (\eg, different cars). In contrast, existing methods can only align 3DGS models of the same object (\eg, the same car) and often must be given true scale as input, while we estimate it successfully. GSA leverages viewpoint-guided spherical map features to obtain robust correspondences and introduces a two-step optimization framework that aligns 3DGS models while keeping them fixed. 
First, we apply an iterative feature-guided absolute orientation solver as our coarse registration, which is robust to poor initialization (\eg, 180° misalignment or a 10× scale gap). Next, we use a fine registration step that enforces multi-view feature consistency, inspired by inverse radiance-field formulations.
The first step already achieves state-of-the-art performance, and the second further improves results. 
In the \emph{same-object} case, GSA outperforms prior works, often by a large margin, even when the other methods are given the true scale. In the harder case of \emph{different objects} in the same category, GSA vastly surpasses them, 
providing the first effective solution for \emph{category-level} 3DGS registration and unlocking new applications. 
Project webpage: \href{https://bgu-cs-vil.github.io/GSA-project/}{https://bgu-cs-vil.github.io/GSA-project}
\end{abstract}

\section{Introduction} \label{Sec:Intro}
3D Gaussian Splatting (3DGS) \cite{kerbl:ACM:2023:3DGS} is 
a powerful representation for efficient, high-fidelity novel view synthesis. However, aligning two independent 3DGS models is an open challenge. 
To date, no dedicated 3DGS registration framework exists, 
let alone at the category level, where inter-model differences (in shape/scale/appearance) complicate the task.
Existing \emph{point cloud} registration methods~\cite{Besl:SensorFusion:1992:ICP,zhou:ECCV:2016:FGR,yew:CVPR:2022:regtr,Qin:CVPR:2022:GeoTransformer} struggle to align 3DGS models, whose point distributions are non-uniform and often have arbitrary scale, position, and orientation as they rely on structure-from-motion (SfM) (\eg, COLMAP \cite{Schoenberger:CVPR:2016:sfm}).

We define \textbf{the 3DGS alignment task} as estimating a \emph{similarity transformation (rotation, translation, and scale)} that aligns one 3DGS model to another, achieving geometric alignment and visual consistency.  
Although also definable at the scene level, we focus on objects.
Moreover, \textbf{we also target the alignment of different objects from the same category} (\eg, cars), where intra-class variation in pose, structure, and detail makes the task especially challenging. 
We introduce \emph{\textbf{G}aussian \textbf{S}platting \textbf{A}lignment (GSA)}, the first method
that solves this task.
This enables applications such as synchronized novel view synthesis across instances, facilitating tasks such as dataset visualization and semantic analysis, as well as object replacement (\autoref{fig:intro_split}, B), where one object replaces another in a geometrically and photometrically consistent manner. This is made possible by precise alignment, and can be seamlessly integrated within 3DGS-based object removal and scene inpainting techniques~\cite{Wang:ECCV:2024:gscream,chen:CVPR:2024:GaussianEditor,huang:Arxiv:2025:3DGSInpainting,kocour:Arxiv:2024:IsThereAnythingLeft}. 

Existing methods, such as GaussReg~\citep{chang:ECCV:2024:gaussreg}, rely on Iterative Closest Point (ICP)~\citep{Besl:SensorFusion:1992:ICP} and struggle to align even models of the \emph{same} object. With  different objects, their performance  collapses due to initialization sensitivity, structural variation, and poor correspondences. In contrast, \emph{GSA} achieves accurate alignment in both cases, outperforming prior methods for same-object registration and yielding the first solution for 3DGS category-level alignment. See \autoref{fig:intro_split}.

GSA starts with building two 3DGS models augmented with \emph{geometry-aware features} that encode spatial and semantic cues for robust correspondence. It then performs coarse registration using a new feature-guided absolute orientation solver to estimate a similarity transformation. Next, it refines the transformation via iterative multi-view feature-consistency optimization.
Unlike prior work, GSA leverages feature-driven reasoning throughout, enabling accurate alignment across models even under large rotations (including 180°), scale variation, and structural noise.

While several 3DGS-based models \cite{Zuo:CVPR:2024:FMGS,Zhou:CVPR:2024:Feature3DGS,Qin:CVPR:2024:LangSplat,Marrie:arXiv:2024:LUDVIG} also lift features (\eg, DINOv2~\citep{Oquab:arXiv:2023:DINOv2}) from 2D to 3D, these features are too ambiguous (as we show) for the alignment task. Thus, when we build the 3DGS models we opt to lift a different type of semantic features, that is geometry-aware~\cite{Mariotti:CVPR:2024:ViewpointSphereMap}. This choice, combined with our method,  enables reliable alignment in challenging conditions.

\textbf{Our key contributions are as follows:} 1)
  We extend ICP to an \emph{iterative absolute-orientation solver} that robustly converges even under poor initializations (\eg, 180° rotations) and large scale mismatches (\eg, 10×). 
  It remains accurate even in cross-instance registration, 
  despite geometric variation, by introducing \emph{semantic-geometric feature constraints} that guide correspondences beyond pure geometry.
  2) A \emph{novel inverse-radiance-field formulation}
   for registration, yielding a \emph{multi-view feature-field consistency method}  that attains near-perfect same-object registration and surpasses prior state-of-the-art by a large margin in cross-instance cases.
 3) Our method enables \textbf{novel applications} including \textbf{geometrically-consistent} object replacement and \textbf{synchronized novel-view synthesis} for coherent rendering and scene manipulation.

\section{Related work}

\textbf{Self-Supervised Visual Features and Geometry-Aware Enhancements.}
Self-supervised methods such as DINO \cite{Caron:ICCV:2021:DINO}, DINOv2 \cite{Oquab:arXiv:2023:DINOv2}, and iBOT \cite{Zhou:ICLR:2022:iBOT} 
yield robust, generalizable visual features. Zhang \etal \cite{Zhang:NeurIPS:2023:TaleOfTwoFeatures} combine DINO with Stable Diffusion for zero-shot correspondence, yet these features still lack 3D geometric awareness.
This motivated recent work to add spatial cues. Zhang~\etal~\cite{Zhang:CVPR:2024:TellingLeftRight} fine-tuned features for geometry-aware matching but struggled with 3D cross-view symmetries.
Mariotti~\etal~\cite{Mariotti:CVPR:2024:ViewpointSphereMap} addressed this by adding weak viewpoint supervision into DINOv2 via spherical maps.
As part of our method, we adopt the 2D features from~\cite{Mariotti:CVPR:2024:ViewpointSphereMap} to improve 3D alignment. 
Unlike prior methods lacking geometric context or struggling with spatial ambiguity, our approach extends 2D semantic alignment principles to 3D, 
enabling spatially-consistent 3DGS registration, even across different objects (within a category).

\textbf{Features in 3DGS.}
Recent works extend 3DGS with semantic features from 2D 
models
for tasks beyond novel view synthesis.
Feature3DGS~\cite{Zhou:CVPR:2024:Feature3DGS} embeds SAM~\cite{Kirillov:ICCV:2023:SAM} and LSeg~\cite{li:Arxiv:2022:LSseg}  via a convolutional decoder for segmentation and editing.
FMGS~\cite{Zuo:CVPR:2024:FMGS} integrates CLIP~\cite{radford:ICML:2021:CLIP} and DINO with multi-resolution hash encoding for efficient detection and segmentation.
LangSplat~\cite{Qin:CVPR:2024:LangSplat} compresses CLIP features into a compact 3D representation and uses SAM to resolve point ambiguities, enabling open-vocabulary queries. While these works target segmentation, editing, or detection, our method advances 3DGS by leveraging geometry-aware features for category-level alignment, enabling consistent alignment across  object instances.

\textbf{Registration in NeRFs and 3DGS.}
The implicit 3D representation of Neural Radiance Fields (NeRF)~\cite{mildenhall:ACM:2021:NeRF} makes registration challenging.
NeRF registration methods include, \eg, DReg-NeRF~\cite{Chen:ICCV:2023:DReg-NeRF}, which  performs voxel-based matching, 
and NeRF2NeRF~\cite{goli:ICRA:2023:nerf2nerf}, which uses annotated keypoints and surface constraints.
Despite progress, NeRF-based registration remains computationally expensive and limited, 
and generalizes poorly across objects.
In contrast, the explicit nature of 3DGS representations~\cite{kerbl:ACM:2023:3DGS} supports direct registration via point-cloud methods.
However, ICP~\cite{Besl:SensorFusion:1992:ICP} requires good initialization, while FGR~\cite{zhou:ECCV:2016:FGR} struggles with noisy or sparse data.
Learning-based methods like REGTR~\cite{yew:CVPR:2022:regtr} predict correspondences using attention but assume rigidity, limiting applicability to 3DGS models with scale variations and uneven point density.
GaussReg~\cite{chang:ECCV:2024:gaussreg} proposes a coarse-to-fine pipeline combining GeoTransformer~\cite{Qin:CVPR:2022:GeoTransformer} with image-guided refinement and explicit handling of 3DGS attributes. 
Existing 3DGS registration works address mainly scene-centric settings (\cite{zhu:3DV:2025:loopsplat,matsuki:CVPR:2024:gsslam,Cheng:2025:ICCV:RegGS,liu:IROS:2025:skeletongsfusion,khatib:ICCVW:2025:gsvisloc}) where the alignment is of different models of the same scene rather than of different objects.
All these methods fail under large-scale variation, 
structural category-level inter-object differences,
and the one-to-many nature of 3DGS (even the same object may yield multiple models).

\textbf{From Inverse Radiance Fields to Cross-Field Registration.}
The inverse radiance-field formulation~\cite{yen:IROS:2021:inerf} estimates a \emph{camera pose} that best explains a target image given a fixed radiance field. While~\cite{yen:IROS:2021:inerf} handled \emph{neural} radiance fields, it was adapted to 3DGS in iComMa~\cite{sun:Arxiv:2023:icomma}.
Building on this idea, \textbf{GSA} generalizes it from single-view camera-pose estimation over the Special Euclidean group $\mathrm{SE}(3)$ (rotation and translation) to multi-view field-to-field registration over the Similarity group $\mathrm{Sim}(3)$ (rotation, translation, and scale).
Rather than optimizing a camera pose to explain an image, we optimize a transformation aligning multi-view renderings across two semantically similar feature fields, possibly at different scales.
This formulation leverages differentiable rendering for direct alignment, enabling accurate same-object and category-level registration.

\begin{algorithm}[t]
\DontPrintSemicolon
    \caption{Preprocessing}
    \label{alg:preprocessing}
    
    \KwIn{
        Original (unmasked) images: \( \widetilde{\Ical}_1 = (\widetilde{I}_1^i)_{i=1}^{N_1} \), \( \widetilde{\Ical}_2 = (\widetilde{I}_2^i)_{i=1}^{N_2} \).
    }
    \KwOut{
        Camera poses: \( \Ccal_1 = (C_1^i)_{i=1}^{N_1} \), \( \Ccal_2 = (C_2^i)_{i=1}^{N_2} \);\\
        Masked images: \( \Ical_1 = (I_1^i)_{i=1}^{N_1} \), \( \Ical_2 = (I_2^i)_{i=1}^{N_2} \);\\
        Masked feature maps: \( \Fcal_1 = (F_1^i)_{i=1}^{N_1} \), \( \Fcal_2 = (F_2^i)_{i=1}^{N_2} \).
    }
    
    Estimate \( \Ccal_1, \Ccal_2 \) using COLMAP~\citep{Schoenberger:CVPR:2016:sfm}\;    
  
    Compute (unmasked) feature maps \( \widetilde{\Fcal}_1 = (\widetilde{F}_1^i)_{i=1}^{N_1} \), \( \widetilde{\Fcal}_2 = (\widetilde{F}_2^i)_{i=1}^{N_2} \)  
    from \( \widetilde{\Ical}_1, \widetilde{\Ical}_2 \) using~\citep{Mariotti:CVPR:2024:ViewpointSphereMap};    

    compute masks \( \Mcal_1, \Mcal_2 \) from \( \widetilde{\Ical}_1, \widetilde{\Ical}_2 \) using~\citep{Kirillov:ICCV:2023:SAM};
    compute masked images: \( \Ical_1 = (I_1^i)_{i=1}^{N_1} \), \( \Ical_2 = (I_2^i)_{i=1}^{N_2} \),  
    where \( I_1^i \) (resp. \( I_2^i \)) is obtained by applying mask \( M_1^i \) (resp. \( M_2^i \)) to \( \widetilde{I}_1^i \) (resp. \( \widetilde{I}_2^i \))

    Compute masked feature maps: \( \Fcal_1 = (F_1^i)_{i=1}^{N_1} \), \( \Fcal_2 = (F_2^i)_{i=1}^{N_2} \),  
    where \( F_1^i \) (resp. \( F_2^i \)) is obtained by applying  \( M_1^i \) (resp. \( M_2^i \)) to \( \widetilde{F}_1^i \) (resp. \( \widetilde{F}_2^i \))
    
    \textbf{Return} \( \Ccal_1, \Ccal_2, \Ical_1, \Ical_2, \Fcal_1, \Fcal_2 \)
\end{algorithm}

\section{Method}
Our goal is to align two independent  \emph{3DGS models}, \( \Gcal_1 \) and \( \Gcal_2 \), using a $\mathrm{Sim}(3)$ transformation. 
These models may represent either the \emph{same object} or, in the harder case, \emph{two different objects} of the same category.  
While prior works on 3DGS alignment can only register \emph{models of the same object}, our method, GSA, is the \emph{first to successfully align 3DGS models of two different objects} within the same category (\eg, two different cars).  We achieve this by 1) leveraging \emph{viewpoint-guided spherical map features} to augment the 3DGS models to facilitate meaningful correspondences and 2) a novel two-step \emph{coarse-to-fine alignment} procedure.  
We now describe our method (outlined in \autoref{Fig:Method}) in detail.

\subimport{./sec/}{fig_method}

\subsection{Preprocessing}\label{Sec:Method:Subsec:Preproceesing}

Our preprocessing, summarized in \autoref{alg:preprocessing}, is as follows. 
Given a set of images depicting an object, we estimate their camera poses using COLMAP~\cite{Schoenberger:CVPR:2016:sfm}
and compute per-image feature maps using~\cite{Mariotti:CVPR:2024:ViewpointSphereMap}. We also extract object masks
using SAM~\cite{Kirillov:ICCV:2023:SAM}.  
These masks are then applied to both the images and the feature maps, producing the masked versions that replace the originals for later stages.  

\begin{algorithm}[t]
\DontPrintSemicolon
    \caption{
    Feature-Augmented 3DGS Models}
    \label{alg:building_3DGS}
    
    \KwIn{
        Masked images: $ \Ical_1 = (I_1^i)_{i=1}^{N_1} $, $ \Ical_2 = (I_2^i)_{i=1}^{N_2} $; 
        Masked feature maps: $ \Fcal_1 = (F_1^i)_{i=1}^{N_1} $, $ \Fcal_2 = (F_2^i)_{i=1}^{N_2} $.
    }
    \KwOut{
        Feature-augmented 3DGS models $ \Gcal_1, \Gcal_2 $.
    }
    
    \tcp{Build $\Gcal_1$ and $\Gcal_2$ using \autoref{Eqn:Loss:Building3DGS}}
    Use $ \Ical_1, \Fcal_1 $ to build $ \Gcal_1 $, lifting colors/features to 3D\;
    Use $ \Ical_2, \Fcal_2 $ to build $ \Gcal_2 $, lifting colors/features to 3D\;    
    \textbf{Return} $ \Gcal_1, \Gcal_2 $
\end{algorithm}

\subsection{Feature-augmented 3DGS}
\label{Sec:Method:Subsec:Aug3DGS}
For each object, we build a feature-enhanced 3DGS model as follows.  
Let $\Ical$, $\Fcal$, and $\Ccal$  
be the collections of masked images, masked feature maps,  
and camera poses associated with the object.  
These were obtained during preprocessing (\autoref{Sec:Method:Subsec:Preproceesing}). 
Henceforth, whenever we refer to images or feature maps, we mean their masked versions.
Let $\Gcal$ be the (feature-augmented) 3DGS model  
we seek to build.  
Each Gaussian \( g \in \Gcal\) is parameterized as
 $   g = (\bp, \bSigma, \alpha, \mathrm{SH}, \boldf)
 $
where \( \bp \in \mathbb{R}^3 \) is the 3D position, \( \bSigma \in \mathbb{R}^{3\times3} \) is the anisotropic covariance matrix, \( \alpha \in [0,1] \) is the opacity,  
$\mathrm{SH}$ represents spherical harmonics (commonly used in 3DGS to model color),  
and  \( \boldf \in \mathbb{R}^3 \) corresponds to the viewpoint-guided spherical-map features~\cite{Mariotti:CVPR:2024:ViewpointSphereMap}.  
That is, while the 3-channel feature maps, $\Fcal$,  
are 2D, each  \( g \) in \( \Gcal \) is assigned a feature vector \( \boldf \), obtained by ``lifting'' the 2D maps in $\Fcal$ into the 3D domain during the construction of the 3DGS model, as described below
and outlined in \autoref{alg:building_3DGS}.

We use the  3DGS~\cite{kerbl:ACM:2023:3DGS} rendering model but in addition to color, we also render features, following~\cite{Zhou:CVPR:2024:Feature3DGS};  
\ie, given a nominal configuration  
of  $\Gcal$ and a camera pose, we render both a color image and a feature map.  
The differentiable rendering enables gradient-based optimization.   
By defining a loss on the rendered images and/or feature maps, we propagate gradients back to the parameters of each  \( g \in \Gcal\).
Concretely, we build $\Gcal$ by minimizing  
a loss,  $\Lcal$,  
which incorporates 
an RGB-based term and 
a feature-based term: 
\begin{align}
    \Lcal &= \Lcal_{\text{rgb}} + \lambda_f \Lcal_f,
    \label{Eqn:Loss:Building3DGS}
\text{ where }
\\ 
    \Lcal_{\text{rgb}} &= (1 - \alpha)  \|I - I^\mathrm{r}\|_{1}    
  + \alpha \Lcal_{\text{SSIM}}(I, I^\mathrm{r})\,,
\\
    \Lcal_f &= \|F - F^\mathrm{r}\|_{1}\,.
\end{align}
Here, $\|\cdot\|_{1}$ is the $\ellOne$ norm,  
$\Lcal_{\text{SSIM}}$ is the Structural Similarity Index Measure (SSIM)~\cite{Wang2003MultiscaleSS},  
$I$ and $F$ are the observed image and feature map, respectively,  
and $I^\mathrm{r}$ and $F^\mathrm{r}$ are their rendered counterparts.
In our experiments,  $\lambda_f=1$ and $\alpha=0.2$.  
The optimization is done using ADAM~\cite{kingma:Arxiv:2014:Adam}.  
\begin{figure}[!t]
\newcommand{\myW}{0.24} 
\centering
\subcaptionbox{}[\myW\linewidth]{
  \includegraphics[trim=30mm 45mm 15mm 45mm, clip, width=\linewidth]{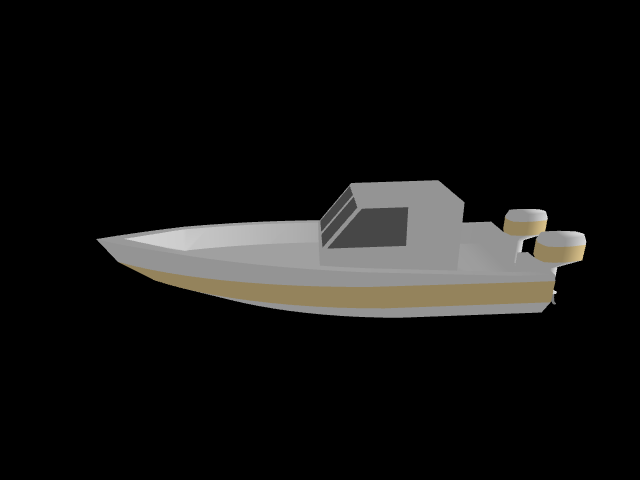}
}
\subcaptionbox{}[\myW\linewidth]{
  \includegraphics[trim=30mm 45mm 15mm 45mm, clip, width=\linewidth]{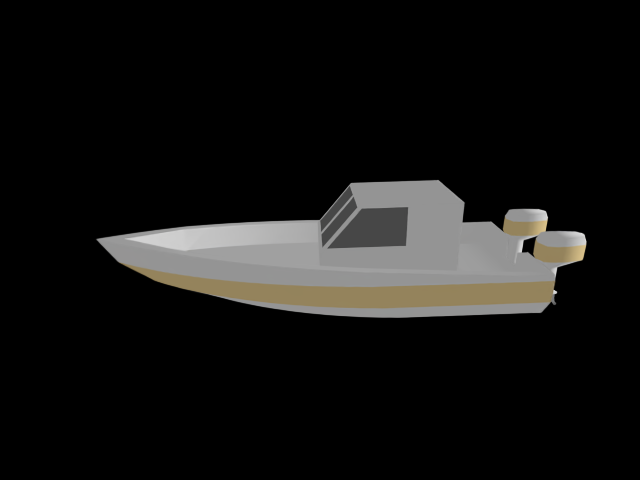}
}
\subcaptionbox{}[\myW\linewidth]{
  \includegraphics[trim=30mm 45mm 15mm 45mm, clip, width=\linewidth]{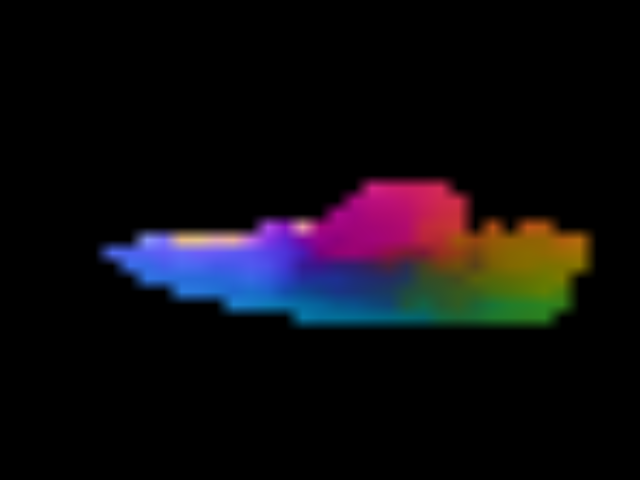}
}
\subcaptionbox{}[\myW\linewidth]{
  \includegraphics[trim=30mm 45mm 15mm 45mm, clip, width=\linewidth]{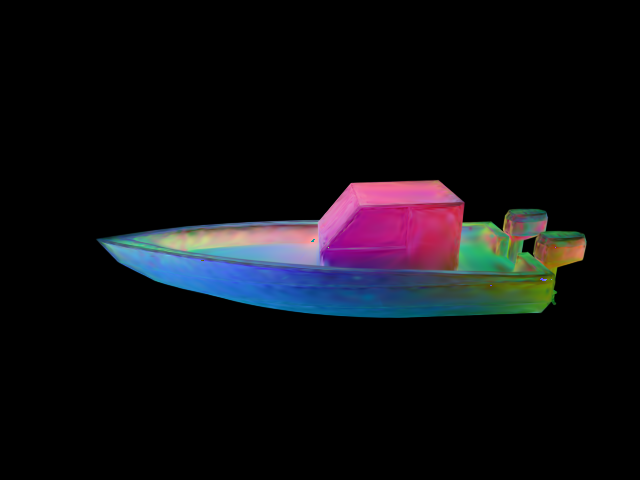}
}
\vspace{-3mm}
\caption{A 3DGS model with enhanced geometric semantic features \cite{Mariotti:CVPR:2024:ViewpointSphereMap}
(a) input image; (b) rendered image;  
(c) low-resolution input feature map;(d) high-resolution rendered feature map.}
\label{fig:enhanced_features}
\end{figure}

Since feature maps' resolution is lower than that of the images, we do not let the feature-based loss affect the geometry.  
That is, and as is done in other methods lifting low-resolution features (\eg, DINO) to 3D~\cite{Qin:CVPR:2024:LangSplat,Zhou:CVPR:2024:Feature3DGS},  
in each iteration of the  construction, we first optimize only over the color and geometry and then fix them while optimizing for lifted features.  
This separation preserves the model's fine geometry while still forcing the lifted features to ``explain away'' the low-resolution 2D feature maps. 
As a result, we render  features at a resolution higher than that of the observed feature maps; see \autoref{fig:enhanced_features}.  
A similar phenomenon was noted in~\cite{Qin:CVPR:2024:LangSplat,Zhou:CVPR:2024:Feature3DGS,yue:ECCV:2024:fit3d}, albeit with different feature types.

We chose the features from~\cite{Mariotti:CVPR:2024:ViewpointSphereMap} over alternatives such as DINOv2~\cite{Oquab:arXiv:2023:DINOv2} or TellingLeftfromRight~\cite{Zhang:CVPR:2024:TellingLeftRight} due to their superior geometry awareness and robustness to spatial ambiguities. 
A detailed comparison between the features, which motivated our choice, appears in our appendix.
\begin{algorithm}[!t]
\DontPrintSemicolon
    \caption{Coarse Alignment
    }
    \label{alg:coarse_registration}
    
    \KwIn{
        Feature-augmented 3DGS models: \( \Gcal_1, \Gcal_2 \).
    }
    \KwOut{
        Similarity transformation \( T = (s, \bR, \bt) \).
    }
    Set initial transformation \( T \) 
    to the identity\;
        \While{the algorithm has not converged and the iteration limit is not exceeded}{
        \ForPar{\( g_i \in T(\Gcal_1)\)}{
            Prune the candidate set in \( \Gcal_2 \) 
            based on feature similarity using the feature means \( \boldf_i \) (\autoref{Eqn:PrunedSet})\;
            Find the closest point (spatially) in the pruned set to 
            the spatial mean, \( \bp_i \) (\autoref{Eqn:ClosestPointFromThePrunedSet})\;
        }
          Minimize \autoref{Eqn:AbsOrientationCostFunction}  
        using a closed-form solution
        (see text) to estimate \( T = (s, \bR, \bt) \)\;
    }    
    \textbf{Return} \( T \)
\end{algorithm}

Lastly, building 3DGS models from masked images with a uniform background often yields  background-colored Gaussians that hurt synthesis; we remove them using a new simple and effective solution; see appendix for details.
\begin{figure}[!t]
  \centering
  \includegraphics[width=1\linewidth, height=0.17\textheight]{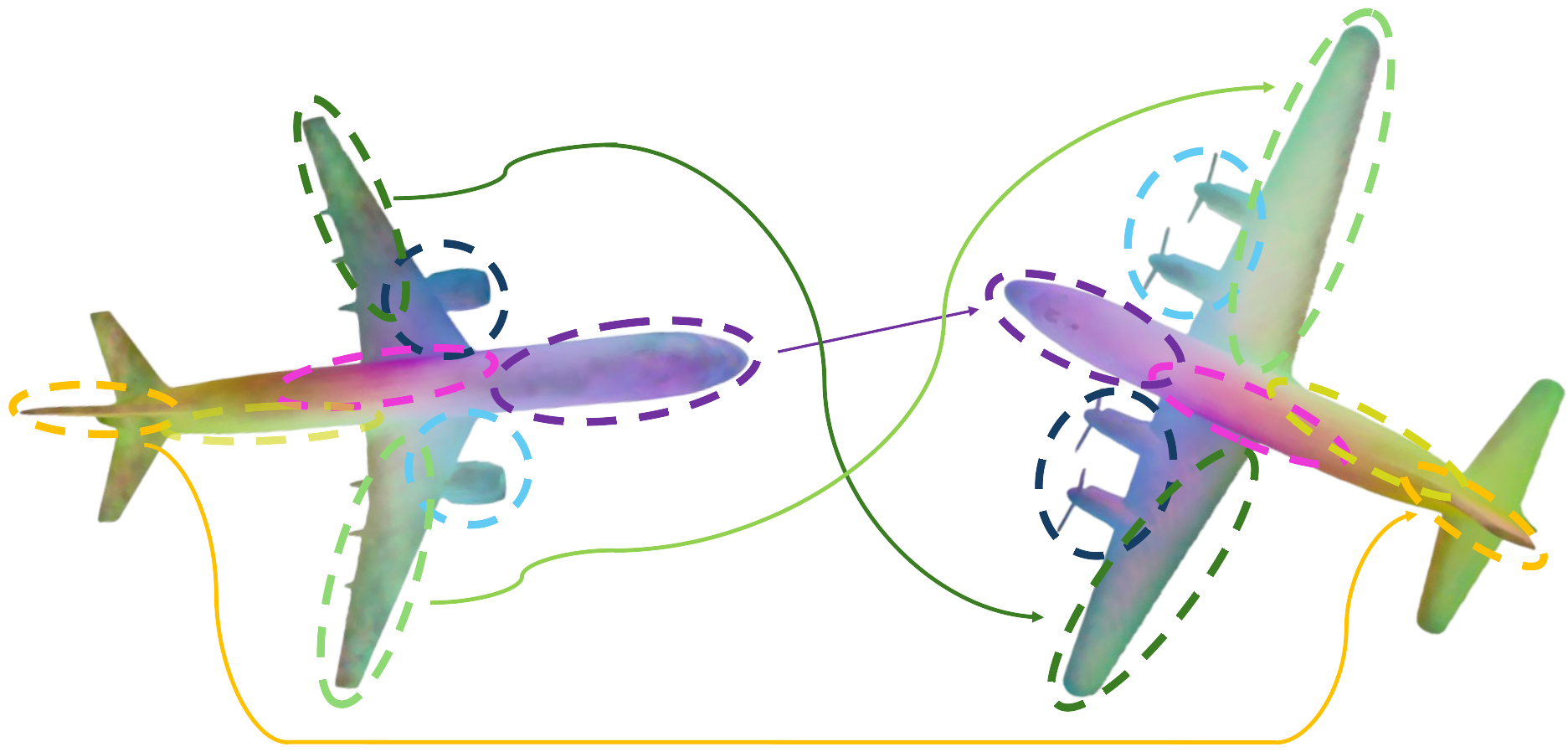}
  \caption{The features from~\cite{Mariotti:CVPR:2024:ViewpointSphereMap} capture both semantic similarity (\eg, wings, engines, tail) and spatial distinctions (\eg, left/right, front/back). Colors indicate matched feature regions across two airplane models; arrows highlight example correspondences.}
  \label{fig:similar_features_regions}
\end{figure}

\subsection{Coarse Alignment}
\label{Sec:Method:Alignment:Coarse}

As outlined in \autoref{alg:coarse_registration}, this step operates entirely in 3D; \ie,  all computations are done on the (fixed) Gaussian representations, without rendering. 
Additionally, by ignoring the spatial covariance matrices and color information, in this step we effectively treat each 3DGS model as a mere point cloud, except that each Gaussian is represented by not only its spatial mean but also feature descriptor.

While our solution for coarse alignment is partially based on ICP~\cite{Besl:SensorFusion:1992:ICP}, 
note that traditional ICP-based methods suffer from three major limitations:
 1)
    They fail under poor initialization (\eg, 180° misalignment) because ICP violates the assumptions of closed-form solvers that require accurate correspondences.
   2)
    They cannot exploit a closed-form solution for similarity transformations (\eg,~\cite{umeyama1991least}): each iteration that alters the scale breaks the geometric consistency of the closest-point correspondences.
    Consequently, they cannot robustly handle an unknown scale and must be given its true value in order to solve for the rotation/translation.
    3) They struggle to find reliable solutions for cross-instance alignment when geometric shapes differ, as the varying geometry causes closest-point matching to converge to incorrect correspondences.
    
By constraining the correspondences using the semantic–geometric feature guidance (see \autoref{fig:similar_features_regions}), our proposed solution goes beyond traditional ICP-based object registration, extending ICP simultaneously in three ways:
1) achieving robustness to extremely poor initializations (including 180° rotation);
2) handling the case of an unknown scale;
3) enabling registration in the cross-instance setting.

Solving for the similarity transformation between two sets of corresponding points is the so-called \emph{absolute orientation problem}, which admits closed-form solutions proposed by Horn~\cite{horn1988closed} and Kabsch–Umeyama~\cite{umeyama1991least}.
We design a hybrid closed-form approach that combines the Kabsch–Umeyama solution for rotation and translation with Horn’s symmetric formulation for scale, which we found to be more reliable for our problem (see Appendix for details).

To estimate the similarity transformation \( T= (s, \bR, \bt) \) between \( \Gcal_1 \) and \( \Gcal_2 \), we iteratively alternate between:  
1) Find correspondences based on spatial proximity. For each point in $ \Gcal_1$, prune the candidate set in the $ \Gcal_2$ by feature similarity.  
2) Given the correspondences, compute the closed-form solution for the optimal transformation.   
3) Apply that transformation to $\Gcal_1$, aligning it toward the $\Gcal_2$.  The process iterates until convergence.
We now provide the details. Let $\bp_i$ and $\bq_j$ denote generic points (\ie, spatial means) in $\Gcal_1$ and $\Gcal_2$, respectively. The similarity transformation is initialized to the identity transformation. Then, in each iteration \( k \), we first transform each \( \bp_i \) in \( \Gcal_1 \) by:  
\begin{align}
\hspace{-2.5mm}
    \bp_i^{(k)} 
    \hspace{-.4mm}=\hspace{-.5mm}
    T^{(k-1)}(\bp_i^{(k-1)})
    \hspace{-.4mm}=\hspace{-.5mm}
    s^{(k-1)}\bR^{(k-1)}\bp_i^{(k-1)}
        \hspace{-.4mm}+\hspace{-.4mm}
    \bt^{(k-1)}
\end{align}
where \( T^{(k-1)}=(s^{(k-1)},\bR^{(k-1)},\bt^{(k-1)}) \) is the transformation from iteration $k-1$.
Next, for each \( \bp_i \), we define a candidate set from \( \Gcal_2 \), based on feature compatibility:
\begin{align}
    \Qcal_i = \{\bq_j \in \Gcal_2 \mid \|\boldf_i - \boldf_j\| \leq \tau_f\}
    \label{Eqn:PrunedSet}
\end{align}
where \( \boldf_i \) and \( \boldf_j \) are the features associated with \( \bp_i \) and \( \bq_j \), respectively. From $\Qcal_i$, we select the spatially closest point:
\begin{align}
    \bq_i^{(k)} = \argmin{\bq \in \Qcal_i} \|\bp_i^{(k)} - \bq\|_{2}.
    \label{Eqn:ClosestPointFromThePrunedSet}
\end{align}
\begin{algorithm}[!t]
\DontPrintSemicolon
    \caption{Fine Alignment}
    \label{alg:fine_registration}  
    \KwIn{
        Feature-augmented 3D Gaussian Splatting models: \( \Gcal_1, \Gcal_2 \)\\
        Initial transformation: \( T = (s, \bR, \bt) \).
    }
    \KwOut{
     Similarity transformation \( T^* = (s, \bR, \bt) \).
    }
    
    \While{the algorithm has not converged and the iteration limit is not exceeded}{
        Render feature maps from \( T(\Gcal_1), \Gcal_2 \)\;
        Compute the multi-view feature consistency loss (\autoref{eq:multi-view-feature-consistency})\;
        Update \( T \) via gradient-based optimization (using ADAM),  
        with gradients propagating through the (fixed) 3DGS representation\;
    }
    
    Set \( T^* \gets T \)
    
    \textbf{Return} \( T^* \)
\end{algorithm}

Given the correspondences \( (\bp_i^{(k)},\bq_i^{(k)}) \), we solve for the optimal  transformation, where optimality is defined via
\begin{align}
  \min_{T^{(k)}\in \mathbf{Sim(3)}} \sum_i \| T^{(k)}(\bp_i^{(k)}) - \bq_i^{(k)}\|_{2}^2\, .
  \label{Eqn:AbsOrientationCostFunction}
\end{align}

Our hybrid closed-form approach, guided by strong geometry-aware semantic features (as shown in \autoref{fig:similar_features_regions}), reliably finds meaningful correspondences and estimates full similarity transformations even in cross-instances, under extremely poor initializations, such as 180° rotations or scale differences of up to 10×, achieving state-of-the-art performance as demonstrated in \autoref{Sec:Results}. Additional qualitative results are included in the appendix.

\subsection{Fine Alignment}
\label{Sec:Method:Alignment:Fine}

We use the coarse estimate from~\autoref{Sec:Method:Alignment:Coarse}
to initialize a finer optimization stage (\autoref{alg:fine_registration}).
This stage enforces consistency between the rendered features of both models across multiple views, ensuring that the estimated transformation improves alignment from diverse perspectives. It is inspired by the inverse-rendering formulation of INeRF~\cite{yen:IROS:2021:inerf}, 
but generalizes the latter from camera space to scene space and extends it to multi-view feature fields.

\begin{table}[t]
\centering
    \centering
    \small

    \begin{tabular}{lcc}
        \toprule
        \textbf{Method} & \textbf{RRE (°) $\downarrow$} & \textbf{ATE $\downarrow$} \\
        \midrule
        FGR \cite{zhou:ECCV:2016:FGR} &  51.32 & 0.33 \\
        REGTR \cite{yew:CVPR:2022:regtr} & 125.34 & 0.38 \\
        DReg-NeRF \cite{Chen:ICCV:2023:DReg-NeRF} & 10.37 & 0.04 \\
        GaussReg \cite{chang:ECCV:2024:gaussreg} & 18.60 & 0.07 \\
       \textbf{(Ours) GSA: Coarse} & 3.13 & 0.01 \\
        \rowcolor{lightyellow} \textbf{(Ours) GSA: Coarse+Fine} & \textbf{0.36} & \textbf{0.001} \\
        \bottomrule
    \end{tabular}
    \caption{Quantitative comparison of registration accuracy for same-object alignment. RRE: Relative Rotation Error (in degrees), ATE: Absolute Translation Error. Lower values are better ($\downarrow$).}
    \label{Table:SameObject}
 \end{table}

Let
$\mathrm{Rend}_{\mathrm{f}}(\Gcal, C)$ denote a differentiable rendering operator that projects the \emph{feature field} of model $\Gcal$ from camera $C$ into a feature map. 
Similarly, let $\mathrm{Rend}_{\mathrm{rgb}}(\Gcal, C)$  denote the analogous rendering of the color radiance field. We now introduce the multi-view feature-consistency loss, which forms the core of our fine alignment stage.
This loss enforces alignment between the rendered feature fields of the two 3DGS models across multiple viewpoints:
\begin{align}
\hspace{-2.15mm}
\label{eq:multi-view-feature-consistency}
\Lcal_{\text{MV-FC}} \hspace{-.15mm}=\hspace{-0.15mm}
\sum_{k=1}^{N}
||
\mathrm{Rend}_{\mathrm{f}}(T\Gcal_1,\,C_k^\star)
\hspace{-.15mm}-\hspace{-0.15mm}
\mathrm{Rend}_{\mathrm{f}}(\Gcal_2,\,C_k^\star)
||_{2}^2
\end{align}
\begin{table*}[h]
    \normalsize
    \centering
    
    \setlength{\tabcolsep}{3.5pt}
    \begin{tabular}{lcccccc}
        \toprule
        \textbf{Method} & \textbf{Airplane} & \textbf{Bus} & \textbf{Boat} & \textbf{Car} & \textbf{Chair} & \textbf{Motorcycle} \\
        & RRE (°) $\downarrow$ & RRE (°) $\downarrow$ & RRE (°) $\downarrow$ & RRE (°) $\downarrow$ & RRE (°) $\downarrow$ & RRE (°) $\downarrow$ \\
        \midrule
        FGR~\cite{zhou:ECCV:2016:FGR} & 90.78 & 94.98 & 86.60 & 89.32 & 91.53 & 82.06 \\
        REGTR~\cite{yew:CVPR:2022:regtr} & 90.00 & 77.69 & 90.98 & 96.98 & 94.75 & 86.97 \\
        GaussReg~\cite{chang:ECCV:2024:gaussreg} & 136.32 & 138.51 & 126.64 & 121.36 & 155.03 & 120.14 \\
        \textbf{GSA: Coarse (Ours)} & 2.97 & 4.74 & 23.97 & 3.70 & 14.17 & \textbf{3.13} \\
        \rowcolor{lightyellow}
        \textbf{GSA: Coarse+Fine (Ours)} & \textbf{2.45} & \textbf{2.41} & \textbf{22.68} & \textbf{1.71} & \textbf{10.86} & 4.05 \\
        \bottomrule
    \end{tabular}
    \caption{Comparison of registration accuracy 
    across categories. Each method is evaluated on six object categories. 
    }
    \label{tab:category_horizontal}
\end{table*}
\renewcommand{\arraystretch}{0.92}
\begin{table}[t]
\centering
\normalsize
\setlength{\tabcolsep}{4pt}
\resizebox{\linewidth}{!}{
\begin{tabular}{llcccc}
\toprule
\textbf{Metric} &
\textbf{Method} &
\textbf{CO3Dv2} &
\textbf{CO3Dv2} &
\textbf{CO3Dv2} &
\textbf{3D Real} \\
 & &
\textbf{toyplane} &
\textbf{bicycle} &
\textbf{chair} &
\textbf{car} \\
\midrule

\multirow{3}{*}{Mean RRE (°) $\downarrow$}
& GaussReg
& 126.09 & 121.12 & 124.45 & 109.72 \\
& GSA (Coarse)
& 7.45
& \textbf{9.19}
& 49.90
& 2.45 \\
& \cellcolor{lightyellow}GSA (Coarse + Fine)
& \cellcolor{lightyellow}\textbf{6.70}
& \cellcolor{lightyellow}13.29
& \cellcolor{lightyellow}\textbf{44.56}
& \cellcolor{lightyellow}\textbf{1.82} \\

\midrule

\multirow{3}{*}{Median RRE (°) $\downarrow$}
& GaussReg
& 121.77 & 122.27 & 125.66 & 107.66 \\
& GSA (Coarse)
& 6.44
& \textbf{4.60}
& 33.12
& 1.93 \\
& \cellcolor{lightyellow}GSA (Coarse + Fine)
& \cellcolor{lightyellow}\textbf{5.22}
& \cellcolor{lightyellow}9.27
& \cellcolor{lightyellow}\textbf{22.67}
& \cellcolor{lightyellow}\textbf{1.34} \\

\bottomrule
\end{tabular}
}
\caption{
RRE (°) ($\downarrow$) on cross-instance alignment of real objects.
}
\label{tab:realworld}
\end{table}

where $||\cdot||_{2}$ is the $\ellTwo$ norm, $T\in \mathrm{Sim}(3)$ is the similarity transformation with scale $s$, rotation $\bR$, and translation $\bt$, $\Gcal_1$ and $\Gcal_2$ are the source and target 3DGS models, and $C_k^\star$ is the $k$-th target camera among the $N$ predefined viewpoints. 
This loss measures the (squared) $\ellTwo$ distance between the two rendered feature maps, and the summation over $k$ enforces consistency across multiple views. 
Intuitively, this loss seeks the transformation $T$ that maximizes multi-view semantic–geometric alignment between the feature renderings of the two models.
The remainder of this section, \emph{which may be safely skipped at first reading}, explains how this formulation arises from generalizing the inverse radiance-field framework to our field-to-field setting.
\begin{figure}[t]
  \centering
\subcaptionbox{\scriptsize FGR}[0.23\linewidth]{
    \includegraphics[width=\linewidth]{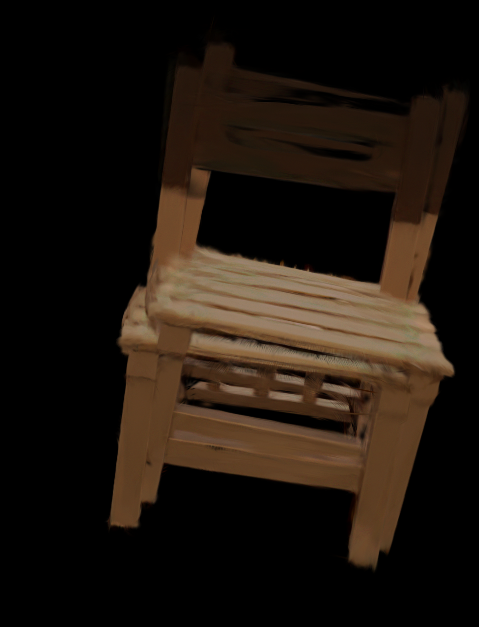}}
  \hfill
  \subcaptionbox{\scriptsize REGTR}[0.23\linewidth]{
    \includegraphics[width=\linewidth]{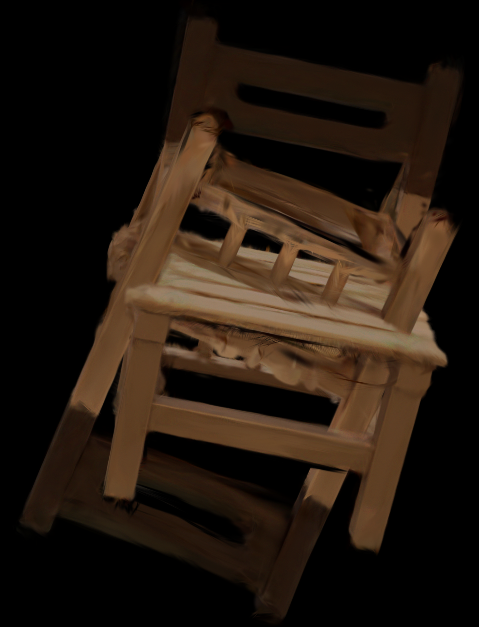}}
  \hfill
  \subcaptionbox{\scriptsize GaussReg}[0.23\linewidth]{
    \includegraphics[width=\linewidth, height=2.51cm]{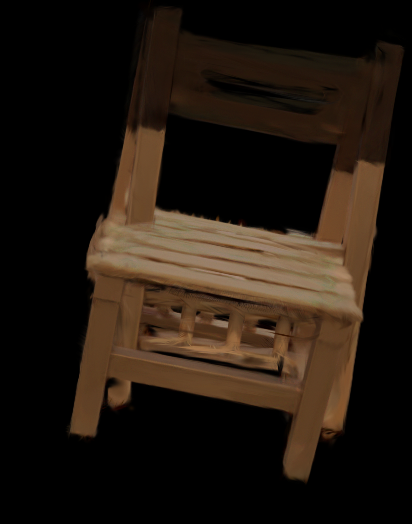}}
  \hfill
  \subcaptionbox{\scriptsize\textbf{GSA}}[0.23\linewidth]{
    \includegraphics[width=\linewidth]{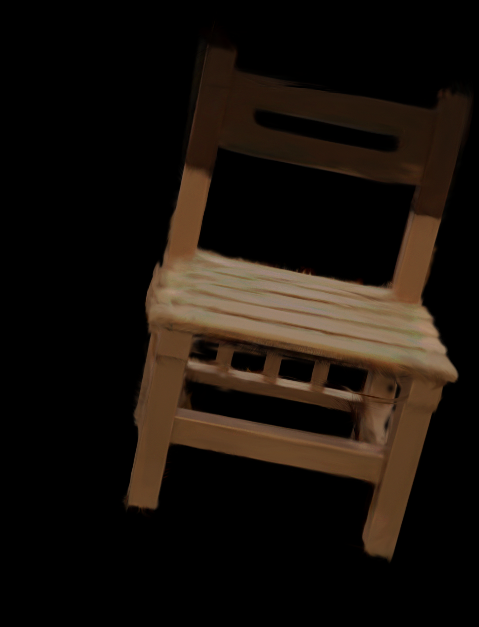}}
  \caption{\textbf{Same-object alignment comparison.}
  Only GSA accurately recovers the full similarity transformation, achieving near-perfect alignment, such that the source and target fully overlap.}
\label{fig:four_examples}
\end{figure}

\textbf{From Inverse NeRF to Scene-Space Optimization.}
In standard NeRF or 3DGS, given camera poses and corresponding images, the goal is to reconstruct a 3D scene such that rendering from the input camera poses reproduces the observed images.
In contrast, the Inverse Radiance Field problem~\cite{yen:IROS:2021:inerf,sun:Arxiv:2023:icomma} takes a known 3D scene $\Gcal$ and a target image $I_t$, and aims to find the camera pose $\Ccal \!\in\! \mathrm{SE}(3)$ that best explains the target image.
Given $I_t$, the pose is estimated 
by $\widehat{\Ccal}^\star$, which is defined
as a solution of 
\begin{align}
\label{eq:inerf-C}
\min_{\Ccal \in \mathrm{SE}(3)}
||\mathrm{Rend}_{\mathrm{rgb}}(\Gcal, \Ccal)- I_t||_{2}^2\,.
\end{align}
Reparameterizing the camera as $\Ccal = T \widehat{\Ccal}_0$, where $\widehat{\Ccal}_0$ is an initial guess and $T\!\in\!\mathrm{SE}(3)$ is a residual motion, yields
\begin{align}
\label{eq:inerf-T}
\min_{T \in \mathrm{SE}(3)}
||\mathrm{Rend}_{\mathrm{rgb}}(\Gcal,
T\widehat{\Ccal}_0)- I_t||_{2}^2\,.
\end{align}
which optimizes the transformation $T$ that produces the camera pose $\Ccal$ explaining $I_t$.
Now, since the rendering depends on the \emph{relative pose} between scene and camera,
\begin{align}
\label{eq:eqv}
\mathrm{Rend}_{\mathrm{rgb}}(\Gcal,\,T \widehat{\Ccal}_0) = \mathrm{Rend}_{\mathrm{rgb}}(T^{-1}\!\Gcal,\,\widehat{\Ccal}_0),
\end{align}
moving the camera by $T$ or the scene by $T^{-1}$ yield identical renderings.
Thus, \autoref{eq:inerf-T} may be rewritten in \emph{scene space}:
\begin{align}
\label{eq:equiv-inerf}
\min_{T \in \mathrm{SE}(3)}
||\mathrm{Rend}_{\mathrm{rgb}}(T^{-1}\!\Gcal,\,\widehat{\Ccal}_0)-I_t||_{2}^2,
\end{align}
which optimizes a rigid transformation of the scene while keeping the initial camera fixed.
We refer to this as the \emph{Scene-Space Inverse Radiance-Field Problem}.

\textbf{Two-field Formulation.}
Let $\Gcal_1$ and $\Gcal_2$ be the source and target radiance fields.
We define the target image as
\begin{align}
I_t = \mathrm{Rend}_{\mathrm{rgb}}(\Gcal_2,\,\Ccal^\star),
\end{align}
where $\Ccal^\star$ is known in the target coordinate frame.
The corresponding optimization becomes
\begin{align}
\hspace{-2mm}
\label{eq:two-field}
\min_{T \in \mathrm{SE}(3)}
||
\mathrm{Rend}_{\mathrm{rgb}}(T\Gcal_1,\,\Ccal_0)-\mathrm{Rend}_{\mathrm{rgb}}(\Gcal_2,\,\Ccal^\star)||_{2}^2.
\end{align}

\textbf{Towards Two-Field Registration.}
To adapt \autoref{eq:two-field} for field-to-field alignment, we assume the existence of a relative transformation between the fields. 
For now, $T^\star\!\in\!\mathrm{SE}(3)$ (later generalized to $\mathrm{Sim}(3)$, since each object might be reconstructed at a different scale as a result of monocular 3D reconstruction).  
Although we cannot assume $\Gcal_2 = T^\star \Gcal_1$ since each model may be built differently (\eg, a different density and number of gaussians) due to separate optimization processes and different training camera-image sets, we can assume that, when rendered through the same volume-rendering function, the two fields produce identical images:
\begin{align}
\hspace{-2mm}
\mathrm{Rend}_{\mathrm{rgb}}(T^\star \Gcal_1, C) = \mathrm{Rend}_{\mathrm{rgb}}(\Gcal_2, C), \quad \forall\, C.
\label{eq:trans_relation}
\end{align}
To recover this unknown $T^\star$, we modify our scene-space inverse rendering objective in \autoref{eq:two-field} by using the same target camera $\Ccal^\star$ instead of a fixed initial guess $\widehat{\Ccal}_0$, yielding
\begin{align}
\hspace{-2mm}
\label{eq:equiv-two-field}
\min_{T \in \mathrm{SE}(3)}
||\mathrm{Rend}_{\mathrm{rgb}}(T\Gcal_1,\,\Ccal^\star)-\mathrm{Rend}_{\mathrm{rgb}}(\Gcal_2,\,\Ccal^\star)||_{2}^2.
\end{align}

\textbf{Extension to Multi-View and $\mathbf{Sim(3)}$.}
The equivalence $\mathrm{Rend}_{\mathrm{rgb}}(\Gcal,\,TC)=\mathrm{Rend}_{\mathrm{rgb}}(T^{-1}\!\Gcal,\,C)$ also holds for $T\!\in\!\mathrm{Sim}(3)$, and thus extends to our formulation.
However, when naively extending \autoref{eq:equiv-two-field} to $\mathrm{Sim}(3)$, with only a single view, a global scale can be fully compensated by an inverse change in camera–scene distance, creating an inherent scale–depth ambiguity.
Introducing multiple target cameras $\{C_k^\star\}_{k=1}^{N}$ jointly constrains all views, thereby eliminating this degeneracy.
The resulting \textbf{Multi-View Radiance Consistency} objective, over $\mathrm{Sim}(3)$, is
\begin{align}
\hspace{-2mm}
\label{eq:final-sim3}
 \min_{T \in \mathrm{Sim}(3)}
\sum_{k=1}^{N}
||
\mathrm{Rend}_{\mathrm{rgb}}(T \Gcal_1,\,C_k^\star)\hspace{-.3mm}-\hspace{-.3mm}
\mathrm{Rend}_{\mathrm{rgb}}(\Gcal_2,\,C_k^\star)
||_{2}^2\, .
\raisetag{3pt}
\end{align}

\textbf{Extension to Feature Fields.}
While no similarity transformation can yield identical renderings between cross-instance radiance fields, leveraging learned semantic–geometric features that are consistent across instances within the same category encourages transformations that maximize multi-view feature consistency, thereby facilitating cross-instance registration, a feat impossible with color radiance fields, since the objects appear visually different.
We formulate this by replacing color rendering ($\mathrm{Render}_{\mathrm{rgb}}$) with feature  rendering ($\mathrm{Render}_{\mathrm{f}}$) which optimizes a single global similarity $T$ that aligns the rendered feature fields across all views.
This defines our fine alignment \textbf{Multi-View Feature-Consistency} loss (\autoref{eq:multi-view-feature-consistency}) in \autoref{alg:fine_registration}.

Unlike previous works~\cite{chang:ECCV:2024:gaussreg,yew:CVPR:2022:regtr,zhou:ECCV:2016:FGR}, which rely on geometric or implicit feature correspondences, we directly exploit the rendering function, under which a true transformation exists for identical objects.  
Therefore, attaining the optimum of our multi-view feature-field consistency objective should theoretically yield \emph{perfect} alignment in the same-object setting, a result confirmed experimentally in~\autoref{Sec:Results:Subsec:SameObject}, where GSA achieves near-perfect registration accuracy.  
In the cross-instance setting~(\autoref{Sec:Results:Subsec:Category}), the same formulation \emph{encourages} transformations that maximize rendering consistency across instances, leading to strong category-level alignment despite inherent structural variation.
\begin{table*}[!h]
    \centering
    \small
   
    \setlength{\tabcolsep}{1pt}
    \begin{tabular}{lcccccc}
        \toprule
        \textbf{Configuration} & \textbf{Airplane} & \textbf{Bus} & \textbf{Boat} & \textbf{Car} & \textbf{Chair} & \textbf{Motorcycle} \\
        & RRE (°) $\downarrow$ & RRE (°) $\downarrow$ & RRE (°) $\downarrow$ & RRE (°) $\downarrow$ & RRE (°) $\downarrow$ & RRE (°) $\downarrow$ \\
        \midrule
        GSA (only coarse) - 1 iteration & 91.04 & 86.79 & 83.10 & 95.52 & 84.42 & 82.52 \\
        GSA (only coarse) - 3 iterations & 3.05 &  4.30 & \textbf{25.17} & 3.27 & 14.54 & \textbf{2.81} \\
        GSA (only coarse) - Until convergence & \textbf{2.97} & \textbf{4.06} & 30.21 & \textbf{3.26} & \textbf{10.50} & 2.89 \\
        \midrule
       GSA (coarse 3 iterations  + fine) - 3 (similar) views &  4.23 & 4.56 & 29.10 & 5.15 & 11.54 & 6.05 \\
        GSA (coarse 3 iterations + fine) - 3 (diverse) views & 2.72 & 3.02 & \textbf{22.68} & \textbf{1.88} & 10.26 & \textbf{3.94} \\
        GSA (coarse 3 iterations  + fine) - 30 views & \textbf{2.41} & \textbf{2.50} & 22.90 & 1.90 & \textbf{5.97} & 4.44 \\
        \midrule
        GSA (coarse 3 iterations + fine) - fine uses colors &  3.11 & 8.64 & 24.95 & 4.95 & 16.06 & 4.80 \\
        \bottomrule
    \end{tabular}
     \caption{
    Testing various GSA configurations 
    on different object categories. 
    Lower is better ($\downarrow$).}
    \label{tab:ours_ablation}

\end{table*}

\section{Experiments and Results}
\label{Sec:Results}

We evaluate \textsc{GSA} on several 3D registration benchmarks and compare it to key registration methods across both Point Cloud and Novel View Synthesis domains. \autoref{subsec:setup} outlines the setup and baselines, \autoref{Sec:Results:Subsec:SameObject} examines same-object registration, and \autoref{Sec:Results:Subsec:Category} extends to category-level cases. \autoref{subsec:real_world} presents real-world results. \autoref{Sec:Results:AblationAndConfig} offers configuration analysis and an ablation study. The appendix shows additional experiments that demonstrate the high accuracy of scale estimation (when the true scale is far from 1) GSA achieves.

\subsection{Experimental Setup}
\label{subsec:setup}
We evaluate GSA under a unified setup.
Coarse registration uses $\tau_f = 0.01$ for up to 6 iterations, and fine registration uses 3 diverse views, 60 iterations, and a learning rate of 0.01.
All experiments were run on an NVIDIA RTX 3090 GPU with identical initialization and view sets for fairness.

\subsection{Aligning two 3DGS Models of the Same Object}
\label{Sec:Results:Subsec:SameObject}
We use the evaluation protocol from~\citep{Chen:ICCV:2023:DReg-NeRF,chang:ECCV:2024:gaussreg} on 15 synthetic objects from Objaverse~\cite{Deitke:CVPR:2023:Objaverse}.
GaussReg and DReg-NeRF assume a fixed scale ($s = 1$), which is generally invalid in Novel View Synthesis since SfM assigns arbitrary scales to point clouds.
Thus, \eg, our main 3DGS competitor, GaussReg, suffers a significant accuracy drop when scale is unknown. To mitigate this, its evaluation assumes a unit scale ($s = 1$), reducing the problem to rigid transformation, a simplification also used by DReg-NeRF, REGTR, and FGR. 
To facilitate a comparison, we ensured the fixed-scale assumption ($s = 1$) holds across all model pairs.  
\textsc{GSA}, without this information, successfully inferred a scale of $\approx 1$, demonstrating robustness to scale ambiguity.
As \autoref{Table:SameObject} shows, GSA substantially improves alignment, achieving an \emph{order-of-magnitude gain} over competing methods.
Notably, our \emph{coarse stage alone} reaches SOTA performance, while the \emph{fine stage} further refines results.

\subsection{Aligning Different Objects Within a Category}
\label{Sec:Results:Subsec:Category}

We evaluate GSA on six ShapeNet~\cite{chang:Arxiv:2015:Shapenet} categories with ground-truth alignment. For each category, we randomly selected five objects and generated images, masks, and camera poses. Within each category, we considered all $\binom{5}{2}=10$ object pairs, yielding 60 pairs overall.
We apply random similarity transforms to each source model, with rotations up to 180° per axis, scales up to $10\times$, and arbitrary translations. These settings test robustness under challenging conditions.
In category-level settings, alignment is a one-to-many problem because intra-class variation leads to ambiguity in scale and translation. We therefore report only the Relative Rotation Error (RRE), a consistent metric across instances. Qualitative results, including scale and translation, appear in \autoref{fig:intro_split}(A) and the appendix.
\autoref{tab:category_horizontal} summarizes the results. Prior methods struggle with cross-instance registration, whereas GSA  attains the lowest error, demonstrating its ability to handle intra-category variation.

\subsection{In-the-wild Within-category Alignment}
\label{subsec:real_world}
After the warm-up (\autoref{Sec:Results:Subsec:SameObject}) and the more challenging but still-synthetic case (\autoref{Sec:Results:Subsec:Category}), we shift to within-category alignment of real-world objects. 
{For evaluation, we manually aligned 5 models from 3D Real Car~\cite{du:Arxiv:2024:3DRealCar} and 10 models from each of three categories in CO3Dv2~\cite{reizenstein:ICCV:2021:CO3D}, yielding $\binom{5}{2} + 3\binom{10}{2} = 145$
test pairs.
We avoided using any privileged dataset information, such as masks or camera poses, and ran the pipeline from scratch. The datasets capture real-world conditions, including partial observations and low-resolution, blurry images, making 
the task
even more difficult.}
As \autoref{tab:realworld} shows, GSA significantly outperforms {the leading 3DGS baseline \cite{chang:ECCV:2024:gaussreg}}. We attribute this success to both feature-based guidance and superior scale estimation. Qualitative results appear in \autoref{fig:intro_split}(B) and the appendix.

\subsection{Configuration Analysis and Ablation Study}\label{Sec:Results:AblationAndConfig}
We begin by analyzing iteration limits and multi-view configurations.
\autoref{tab:ours_ablation} shows the effect of more coarse-step iterations and additional fine-step views. The results show that even with only 3 coarse-step iterations and 3 diverse fine-step views, accuracy remains on par with configurations using more iterations or views. 
Thus, increasing these numbers gives no benefit. 
Of note, using 3 too similar views did hurt  results. 
In fine alignment, we also tested replacing feature rendering with standard color rendering,
as \autoref{tab:ours_ablation}'s last row shows, this caused a significant accuracy drop.

As explained in the appendix, using features from ~\cite{Oquab:arXiv:2023:DINOv2} or ~\cite{Zhang:CVPR:2024:TellingLeftRight} (instead of ~\cite{Mariotti:CVPR:2024:ViewpointSphereMap}) usually \emph{completely} fails, making quantitative comparison pointless; we illustrate this qualitatively in the same section. The appendix also shows the effect of keeping background Gaussians.
Finally, we tested dropping feature-based guidance in coarse alignment; \ie, disabling pruning and basing correspondences solely on spatial proximity (as in regular ICP).
Re-evaluation on Objaverse showed a drastic performance drop: coarse alignment RRE rose to 136.29$^\circ$, and fine alignment to 139.82$^\circ$.

\section{Conclusion}\label{Sec:Conclusion}
We presented GSA, a novel method for registering 3DGS models. Unlike prior work, which fails to align 3DGS models across different objects, GSA achieves robust category-level registration. 
GSA significantly outperforms existing methods, particularly in cross-object scenarios, and integrates seamlessly into standard 3DGS pipelines. 
Our appendix includes additional visual results and applications. 
GSA's main \textbf{limitation} is that its performance depends on the quality of the geometry-aware features. If these
are suboptimal, alignment accuracy may degrade. However, geometry-aware features are a rapidly evolving field, and advances in this area will behoove GSA. 

\clearpage
\FloatBarrier

\section*{Acknowledgments}
This work was supported by the Lynn and William Frankel Center at BGU CS. Roy Amoyal was also supported by the Kreitman School of Advanced Graduate Studies.
\bibliographystyle{ieeenat_fullname}

\bibliography{refs}

\clearpage
\onecolumn
\appendix
\setcounter{page}{1}
\renewcommand{\thepage}{S\arabic{page}}

\begin{center}
    {\LARGE Cross-Instance Gaussian Splatting Registration via\par}
    \vspace{0.3em}
    {\LARGE Geometry-Aware Feature-Guided Alignment\par}
    \vspace{0.5em}
    {\large ------------\par}
    \vspace{0.5em}
    {\LARGE Supplemental Material\par}
\end{center}

\vspace{1cm}

\renewcommand{\sectionautorefname}{Appendix}
\renewcommand{\subsectionautorefname}{Appendix}

\renewcommand{\thefigure}{A.\arabic{figure}} 
\setcounter{figure}{0} 

\noindent\textbf{Appendix Contents:}
    \begin{itemize}
      \item \autoref{appendix:additional_visual_results} - Additional Visual Results
      \item \autoref{appendix:applications} - Applications
      \item \autoref{appendix:scale_estimation} - Scale Estimation Results
      \item \autoref{appendix:features_impact} - The Impact of Feature Choice on Alignment
      \item \autoref{appendix:black_gaussians} - Background Gaussians in 3DGS and Our Solution
      \item \autoref{appendix:complexity_runtime} - Computational Complexity and Runtime Analysis
      \item \autoref{appendix:closed_form_solutions} - Hybrid Kabsch-Umeyama and Horn Closed-Form Solution to the Absolute Orientation Problem
    \end{itemize}

\clearpage
\section{Additional Visual Results}
\label{appendix:additional_visual_results}

\begin{figure}[h]
    \centering
    \newcommand{\myImgW}{0.17\textwidth}  

    \makebox[\textwidth][c]{%
        \subcaptionbox{FGR}[\myImgW]{%
            \includegraphics[trim=15mm 0mm 0mm 0mm, clip,
                             width=\myImgW]{./figures/qualitative/fgr.png}
        }\hspace{0.01\textwidth}
        \subcaptionbox{REGTR}[\myImgW]{%
            \includegraphics[trim=15mm 0mm 0mm 0mm, clip,
                             width=\myImgW]{./figures/qualitative/regtr.png}
        }\hspace{0.01\textwidth}
        \subcaptionbox{GaussReg}[\myImgW]{%
            \includegraphics[trim=15mm 0mm 0mm 0mm, clip,
                             width=\myImgW]{./figures/qualitative/gaussreg.png}
        }\hspace{0.01\textwidth}
        \subcaptionbox{\footnotesize \textbf{GSA: coarse (ours)}}[\myImgW]{%
            \includegraphics[trim=15mm 0mm 0mm 0mm, clip,
                             width=\myImgW]{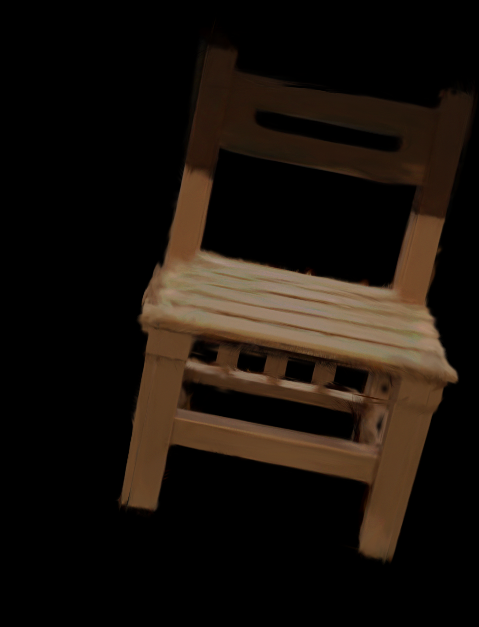}
        }\hspace{0.01\textwidth}
        \subcaptionbox{\footnotesize \textbf{GSA: fine (ours)}}[\myImgW]{%
            \includegraphics[trim=15mm 0mm 0mm 0mm, clip,
                             width=\myImgW]{./figures/qualitative/gsa_coarse_fine.png}
        }
    }

    \caption{A Comparison of alignment results on two 3DGS models of the same chair from the Objaverse dataset. Although the object has a clear and rigid structure, classical and learning-based methods struggle to achieve consistent alignment. Our GSA method, especially with coarse and fine refinement, achieves the most accurate and coherent results.}
    \label{fig:qualitative_chair}
\end{figure}

\autoref{fig:qualitative_chair} 
depicts a case where the task 
is to align two models of the same object, as described in the first experiment in the paper.
In this particular case, it is the same chair. As \autoref{fig:qualitative_chair}
shows, the competing methods
(FGR, REGTR, and GaussReg)
struggle with the geometric structure
of the chair. In contrast, our GSA, 
even when using only its coarse step, achieves almost-perfect results. Adding the fine step  here did improve the result a bit further, but as the coarse step was so successful, the visual difference is too subtle to note.

\begin{figure}[h]
    \centering
    \resizebox{1\textwidth}{!}{ 
        \begin{minipage}{\textwidth}
            \newcommand{\myW}{0.3}  

            \subcaptionbox{FGR}[.3\textwidth]%
            {\includegraphics[trim=0mm 25mm 0mm 0mm, clip, width=\linewidth]{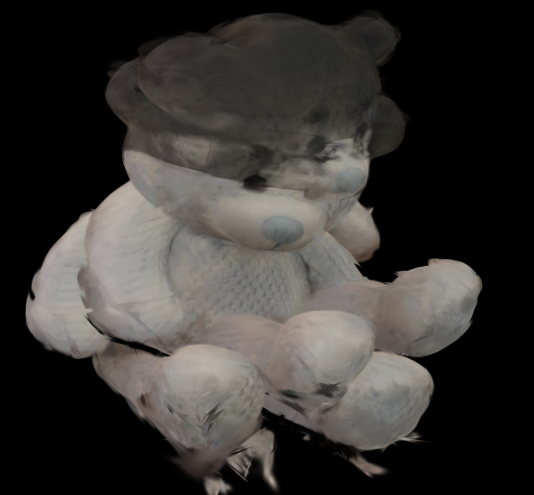}}
            \hfill
            \subcaptionbox{REGTR}[.3\textwidth]%
            {\includegraphics[trim=0mm 25mm 0mm 0mm, clip, width=\linewidth]{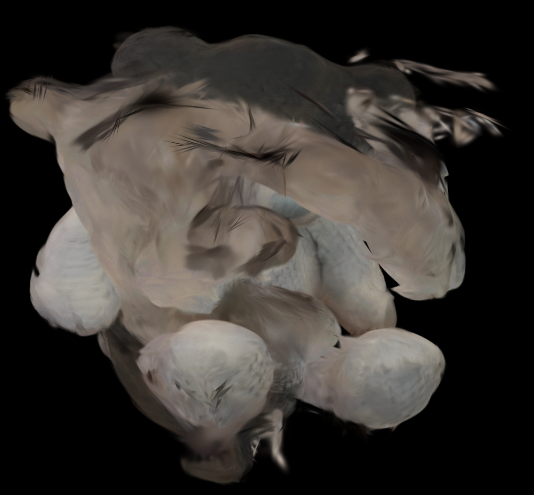}}
            \hfill
            \subcaptionbox{GaussReg}[.3\textwidth]%
            {\includegraphics[trim=0mm 25mm 0mm 0mm, clip, width=\linewidth]{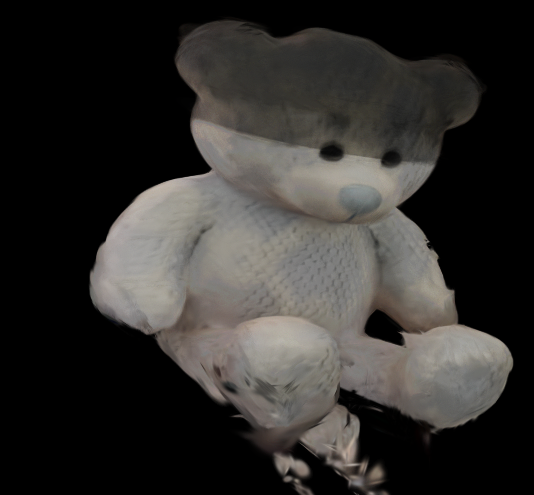}}

            \vspace{0.5em}  

            \hspace{0.15\textwidth} 
            \subcaptionbox{\textbf{GSA: coarse (ours)}}[.3\textwidth]%
            {\includegraphics[trim=0mm 25mm 0mm 0mm, clip, width=\linewidth]{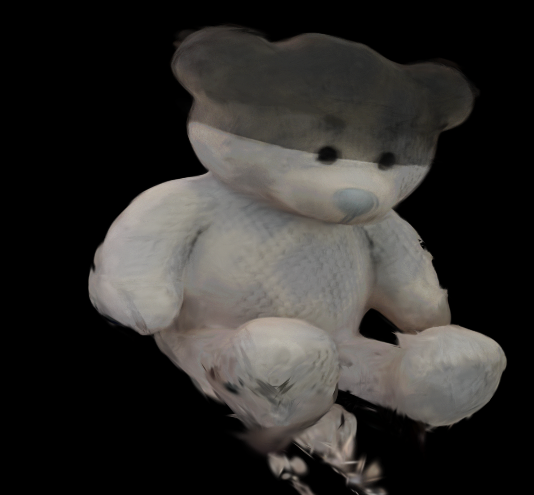}}
            \hfill
            \subcaptionbox{\textbf{GSA: fine (ours)}}[.3\textwidth]%
            {\includegraphics[trim=0mm 25mm 0mm 0mm, clip, width=\linewidth]{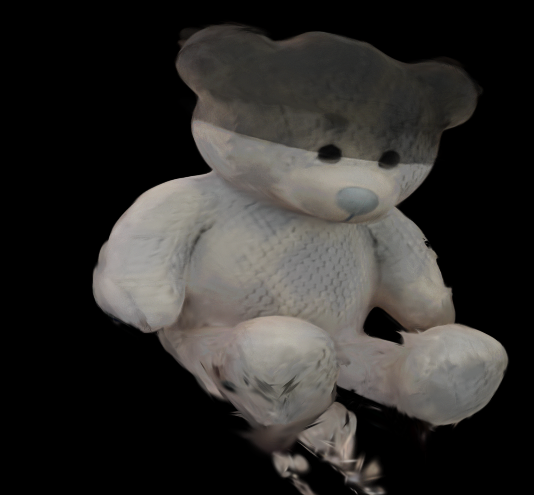}}
            \hspace{0.15\textwidth} 
        \end{minipage}
    }
    \caption{Comparison of two 3DGS models of the same teddy bear, Objaverse~\cite{Deitke:CVPR:2023:Objaverse} dataset. Although the object is somewhat simple and although there is a unique rigid transformation solution, existing point cloud-based registration methods still struggle while our coarse point-cloud registration achieves almost-perfect results (so in this particular case,  the further improvement by our fine step 
    is hard to notice visually).}
    \label{fig:qualitative_teddybear}
\end{figure}

\autoref{fig:qualitative_teddybear}
depicts
another same-object example, 
this time of a teddy bear. 
Again, GSA successfully aligned
the models. 
The cloud-based methods (FGR, REGR) 
struggle here too. 
Here, GaussReg did better than it did with the chair. That said, 
quantitatively, GSA's result
was still better, even if it is
hard to tell this fact by visual inspection.

\begin{figure*}[h]
    \centering

    \hspace{1.4cm}
    \begin{minipage}{0.28\linewidth}\centering \large \textbf{Random Initialization} \end{minipage}%
    \hspace{1pt}
    \begin{minipage}{0.28\linewidth}\centering \large \textbf{GSA: coarse} \end{minipage}%
    \hspace{1pt}
    \begin{minipage}{0.28\linewidth}\centering \large \textbf{GSA: coarse+fine} \end{minipage}

    \vspace{-2pt}

    \noindent
    \begin{minipage}[c][3.2cm][c]{1.4cm}
        \centering \large \textbf{View 1:}
    \end{minipage}%
    \begin{minipage}[c][3.2cm][c]{0.24\linewidth}
        \includegraphics[width=\linewidth]{./figures/qualitative/category/qualitative_scale_results/boats/random_boats_view1.png}
    \end{minipage}%
    \hspace{1pt}
    \begin{minipage}[c][3.2cm][c]{0.24\linewidth}
        \includegraphics[width=\linewidth]{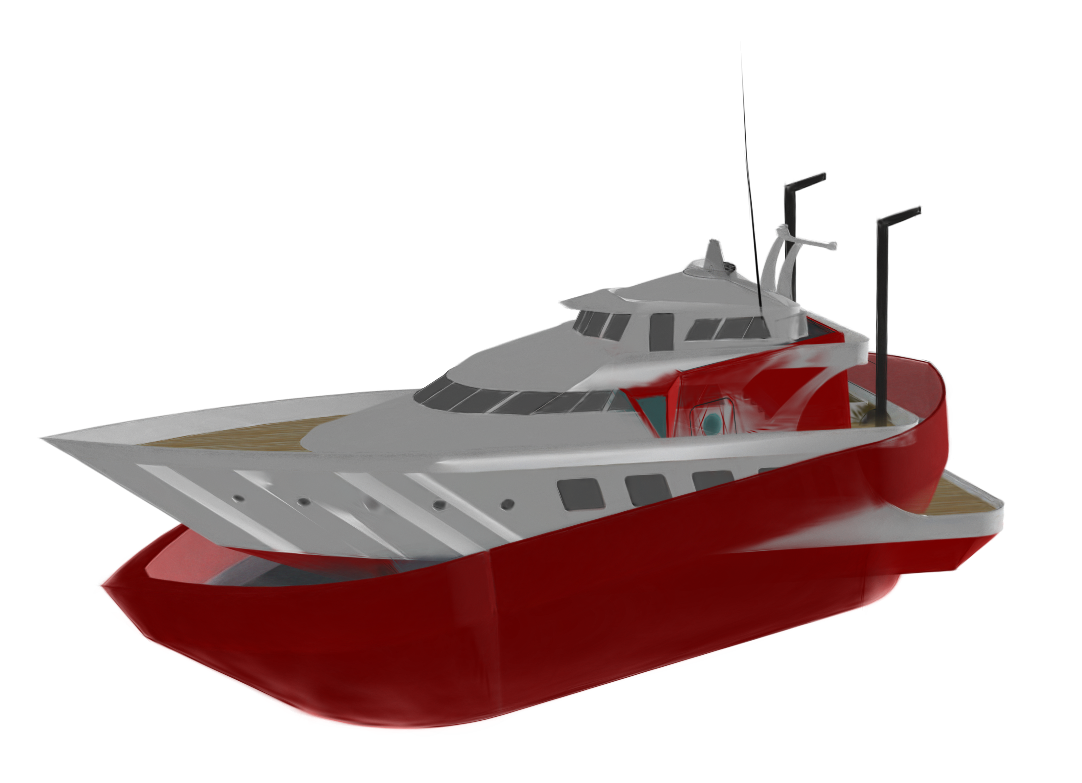}
    \end{minipage}%
    \hspace{1pt}
    \begin{minipage}[c][3.2cm][c]{0.24\linewidth}
        \includegraphics[width=\linewidth]{./figures/qualitative/category/qualitative_scale_results/boats/fine_boats_view1.png}
    \end{minipage}


    \noindent
    \begin{minipage}[c][3.2cm][c]{1.4cm}
        \centering \large \textbf{View 2:}
    \end{minipage}%
    \begin{minipage}[c][3.2cm][c]{0.24\linewidth}
        \includegraphics[width=\linewidth]{./figures/qualitative/category/qualitative_scale_results/boats/random_boats_view2.png}
    \end{minipage}%
    \hspace{1pt}
    \begin{minipage}[c][3.2cm][c]{0.24\linewidth}
        \includegraphics[width=\linewidth]{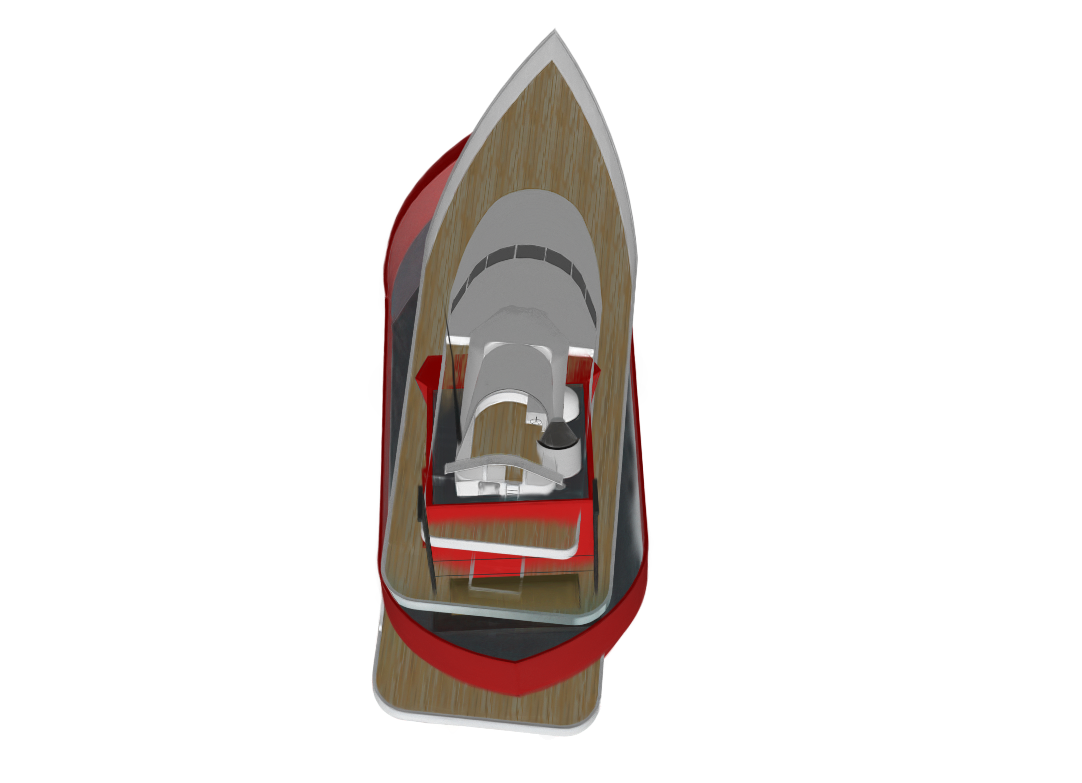}
    \end{minipage}%
    \hspace{1pt}
    \begin{minipage}[c][3.2cm][c]{0.24\linewidth}
        \includegraphics[width=\linewidth]{./figures/qualitative/category/qualitative_scale_results/boats/fine_boats_view2.png}
    \end{minipage}

    \vspace{10pt}
    \noindent\hdashrule[0.5ex]{\linewidth}{1pt}{3pt 2pt}
    
    \noindent
    \begin{minipage}[c][3.2cm][c]{1.4cm}
         \centering \large \textbf{View 1:}
    \end{minipage}%
    \begin{minipage}[c][3.2cm][c]{0.24\linewidth}
        \includegraphics[width=\linewidth]{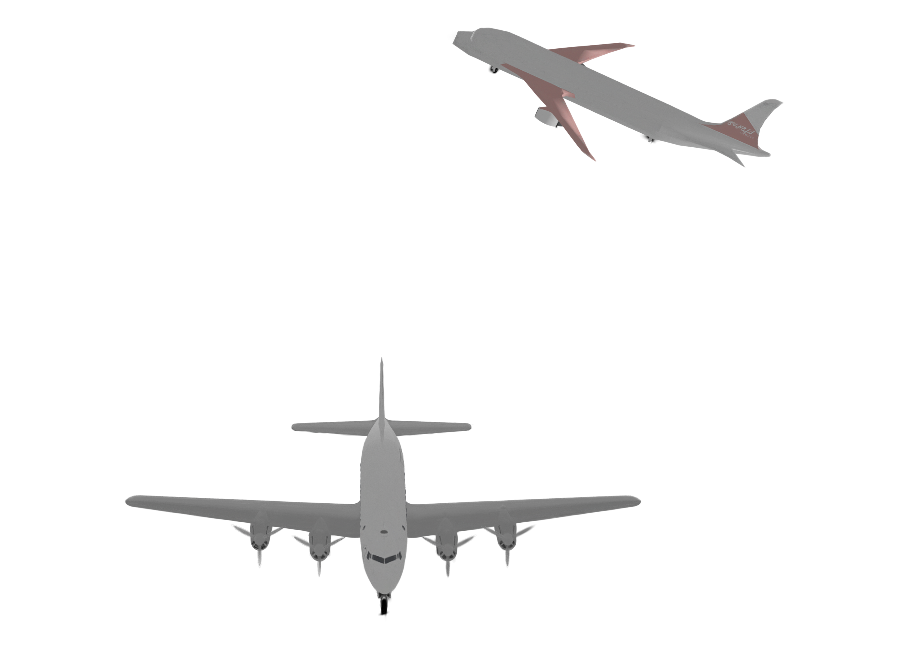}
    \end{minipage}%
    \hspace{1pt}
    \begin{minipage}[c][3.2cm][c]{0.24\linewidth}
        \includegraphics[width=\linewidth]{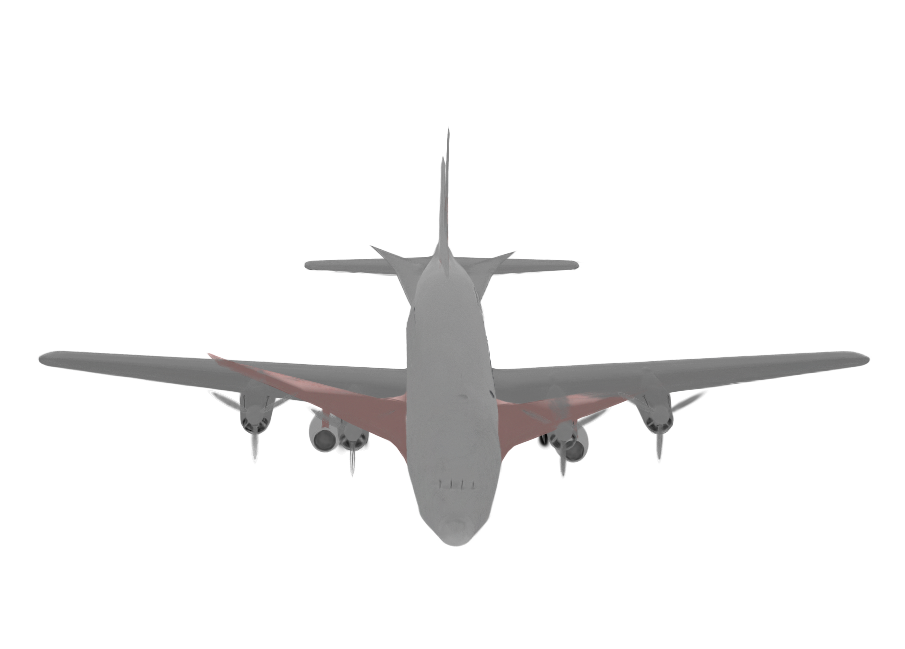}
    \end{minipage}%
    \hspace{1pt}
    \begin{minipage}[c][3.2cm][c]{0.24\linewidth}
        \includegraphics[width=\linewidth]{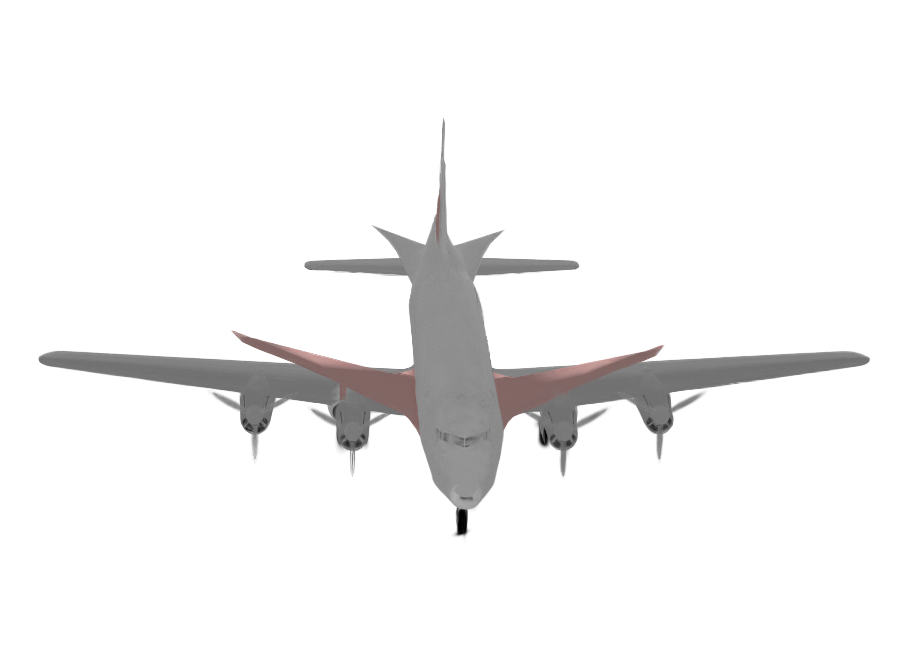}
    \end{minipage}


    \noindent
    \begin{minipage}[c][3.2cm][c]{1.4cm}
        \centering \large \textbf{View 2:}
    \end{minipage}%
    \begin{minipage}[c][3.2cm][c]{0.24\linewidth}
        \includegraphics[width=\linewidth]{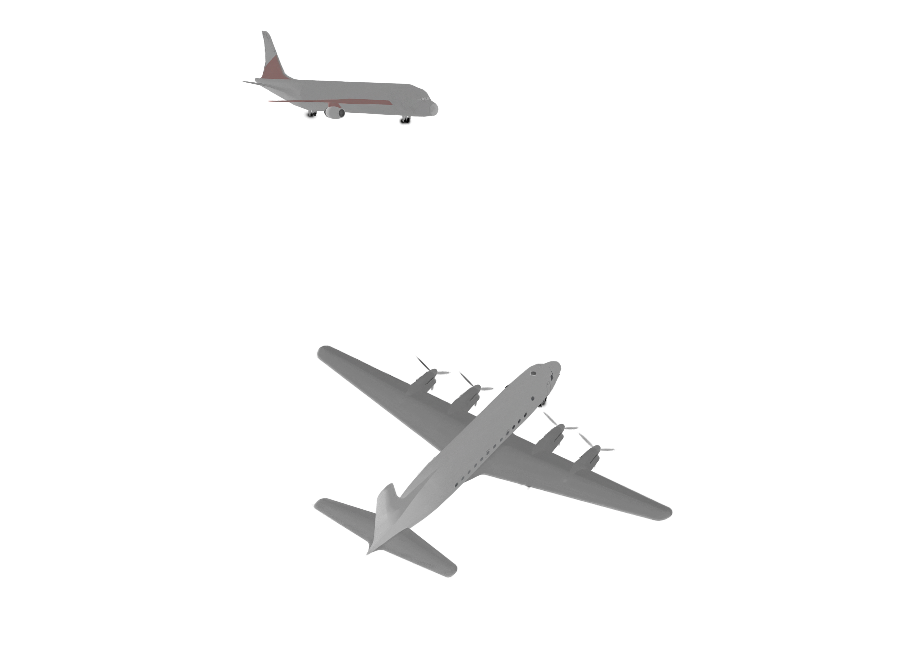}
    \end{minipage}%
    \hspace{1pt}
    \begin{minipage}[c][3.2cm][c]{0.24\linewidth}
        \includegraphics[width=\linewidth]{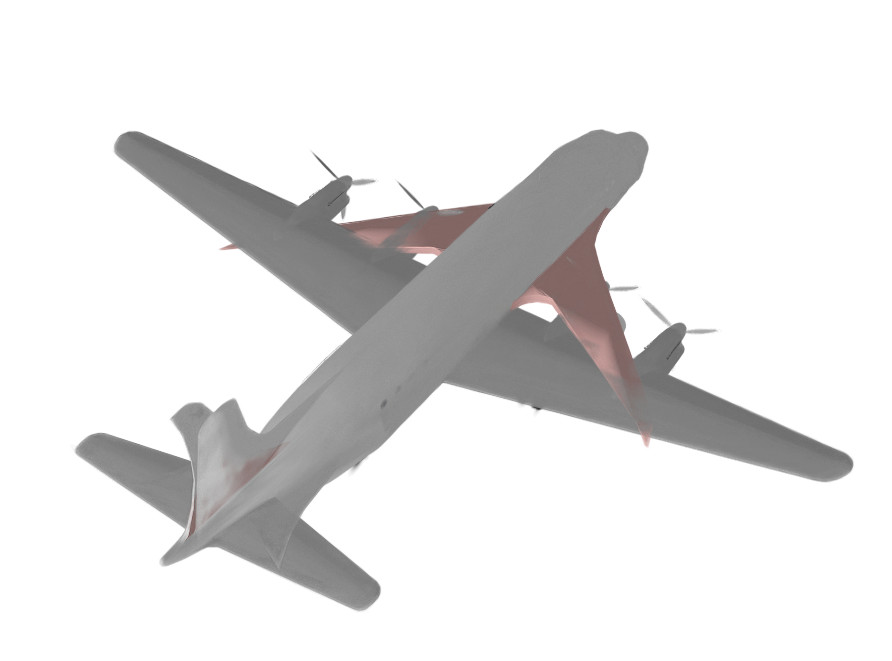}
    \end{minipage}%
    \hspace{1pt}
    \begin{minipage}[c][3.2cm][c]{0.24\linewidth}
        \includegraphics[width=\linewidth]{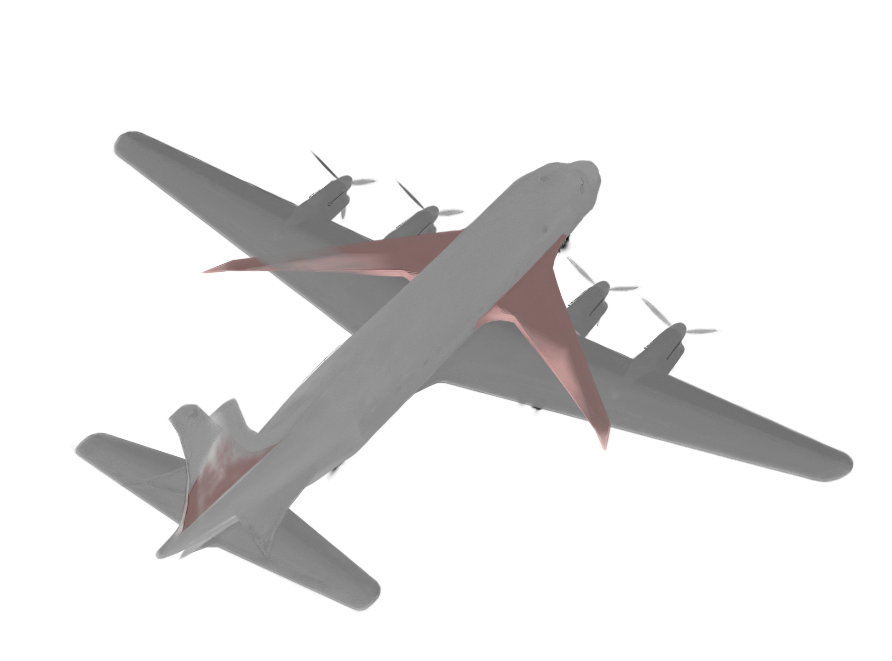}
    \end{minipage}

    \vspace{10pt}
    \noindent\hdashrule[0.5ex]{\linewidth}{1pt}{3pt 2pt}
    \vspace{3pt}

    \noindent
    \begin{minipage}[c][3.2cm][c]{1.4cm}
        \centering \large \textbf{View 1:}
    \end{minipage}%
    \begin{minipage}[c][3.2cm][c]{0.24\linewidth}
        \includegraphics[width=\linewidth]{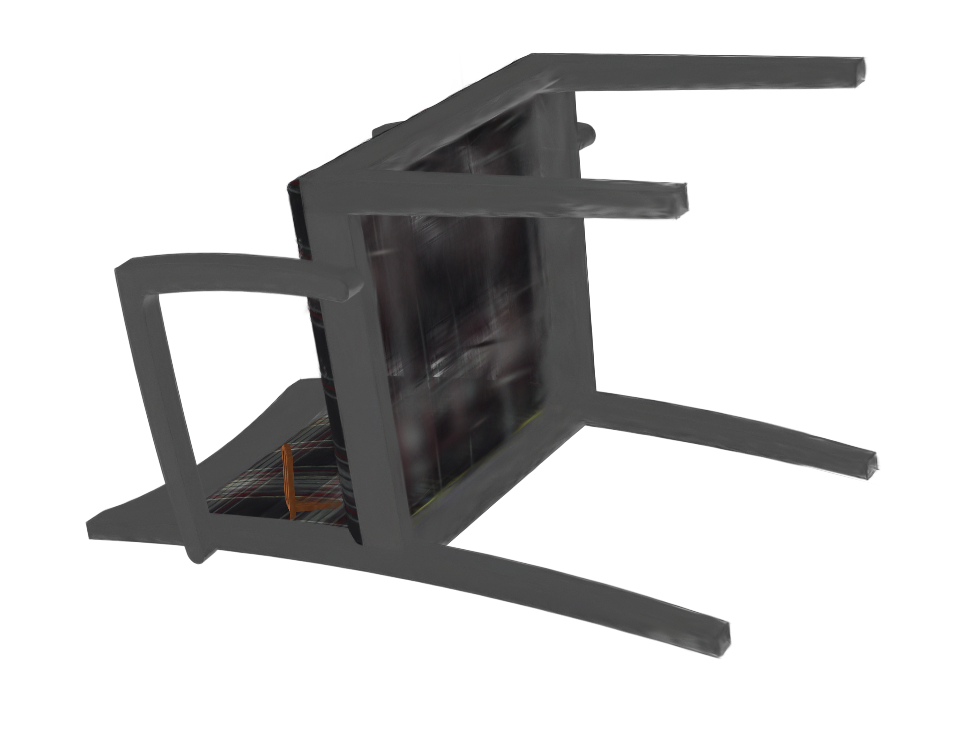}
    \end{minipage}%
    \hspace{1pt}
    \begin{minipage}[c][3.2cm][c]{0.24\linewidth}
        \includegraphics[width=\linewidth]{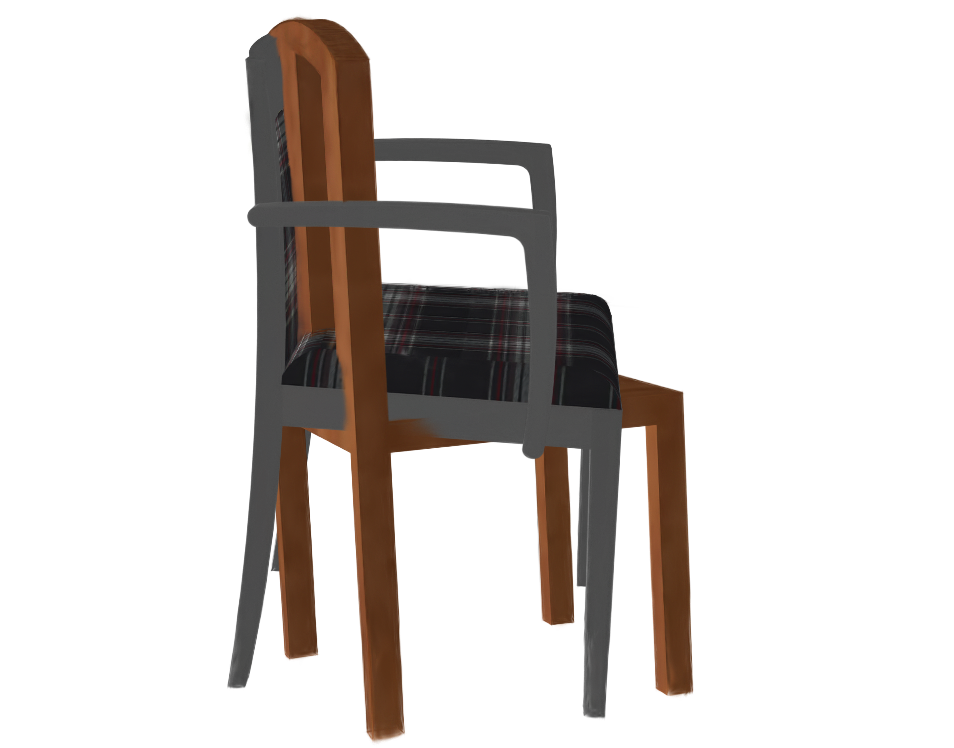}
    \end{minipage}%
    \hspace{1pt}
    \begin{minipage}[c][3.2cm][c]{0.24\linewidth}
        \includegraphics[width=\linewidth]{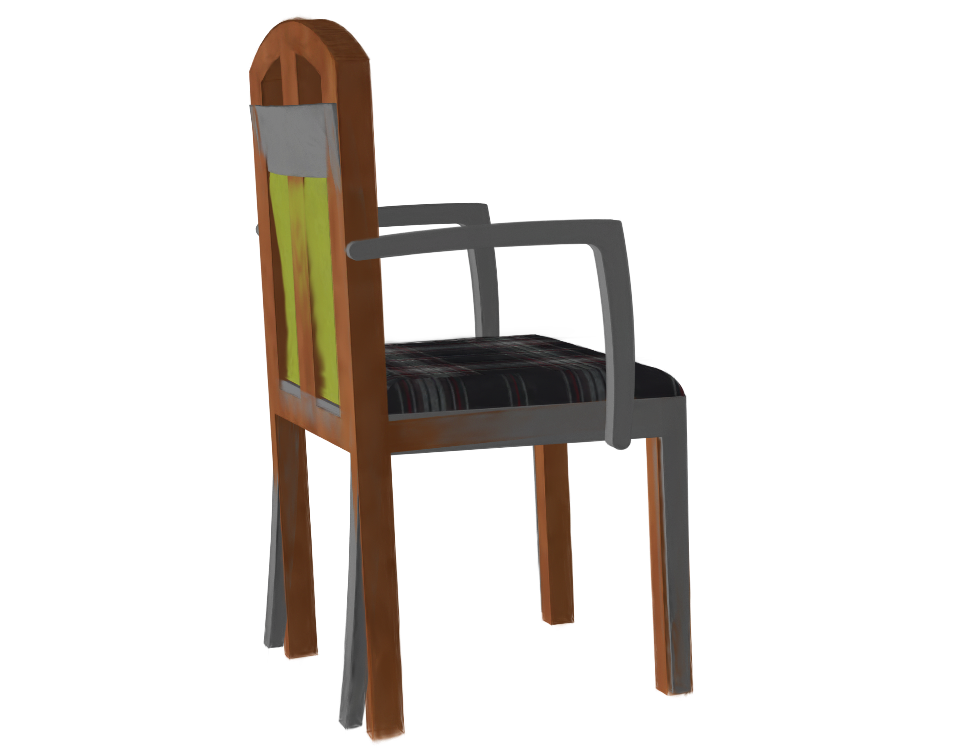}
    \end{minipage}


    \noindent
    \begin{minipage}[c][3.2cm][c]{1.4cm}
        \centering \large \textbf{View 2:}
    \end{minipage}%
    \begin{minipage}[c][3.2cm][c]{0.24\linewidth}
        \includegraphics[width=\linewidth]{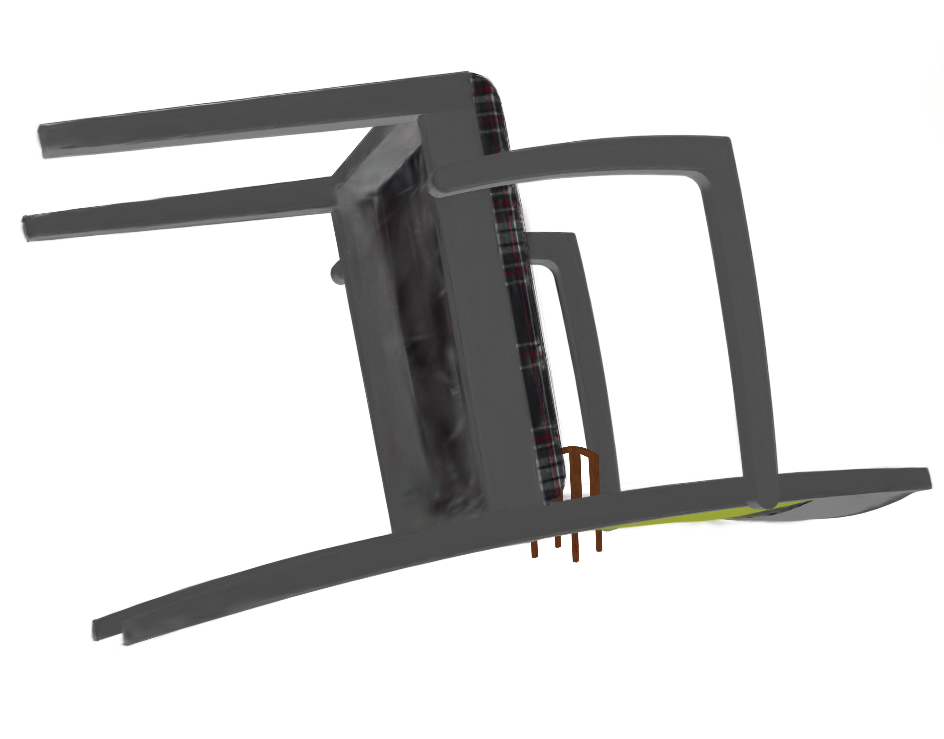}
    \end{minipage}%
    \hspace{1pt}
    \begin{minipage}[c][3.2cm][c]{0.24\linewidth}
        \includegraphics[width=\linewidth]{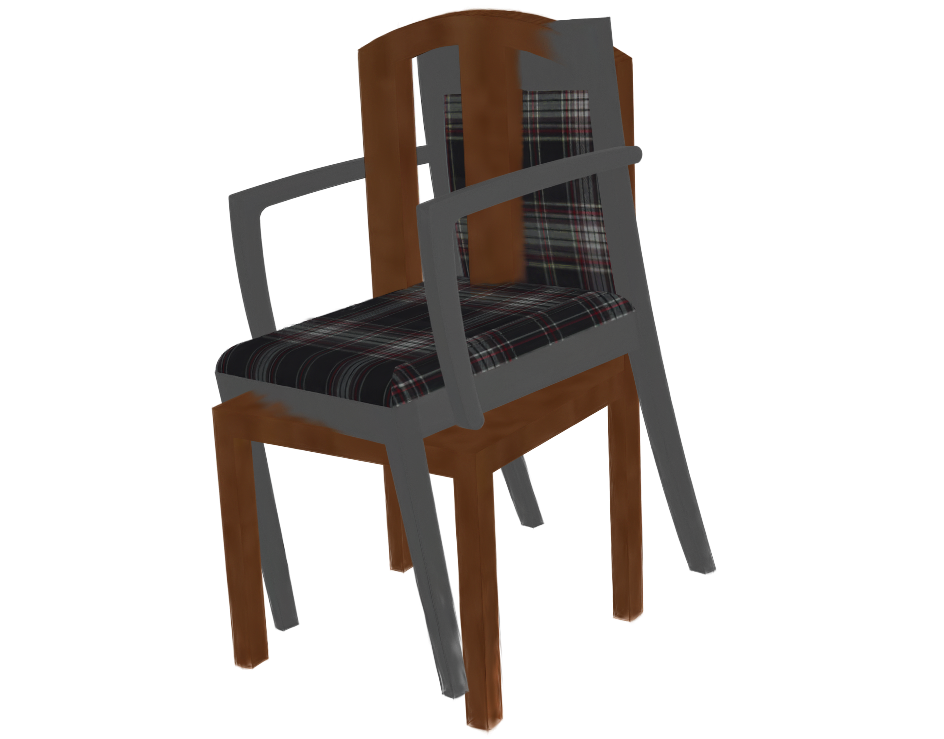}
    \end{minipage}%
    \hspace{1pt}
    \begin{minipage}[c][3.2cm][c]{0.24\linewidth}
        \includegraphics[width=\linewidth]{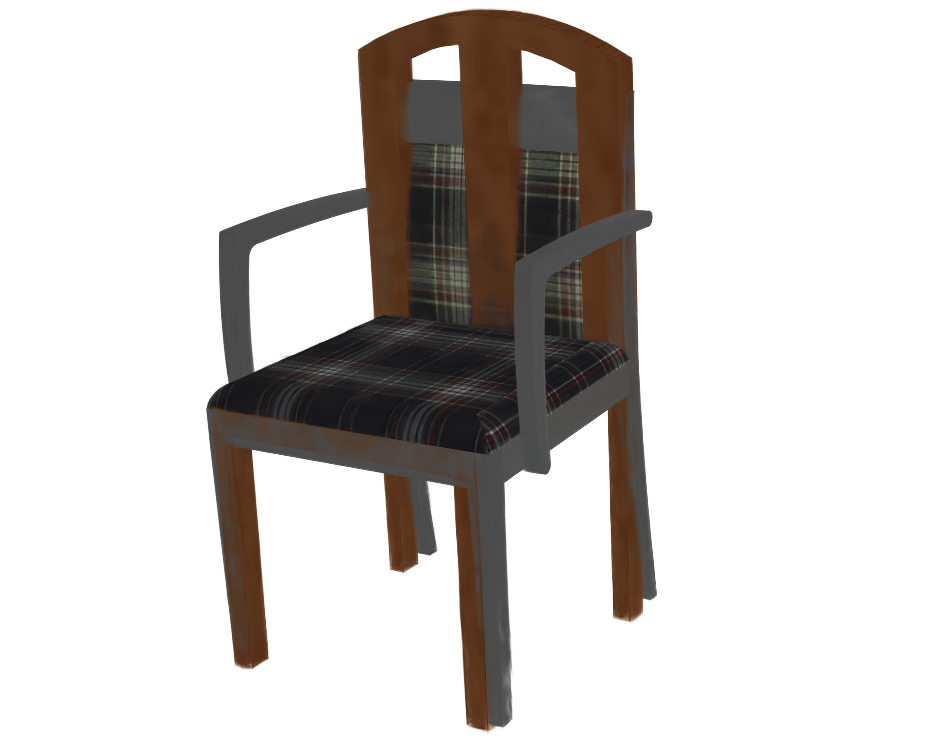}
    \end{minipage}

    \vspace{-5pt}
    
 \caption{Visual qualitative results for aligning
 different objects within the same cateogry. 
 In each of the three 2-row blocks we show an example
 form a different category (boats, airplanes, and chairs). 
 The two rows within each 2-row block show different views.
 In each 2-row block we show, from lest to right:
 1) a random initialization; 2) the result of GSA's coarse step;
 3) the result of GSA's fine step.   
 The random initialization includes arbitrary translations, rotations up to 180 degrees, and significant scale differences.}

\label{fig:qualitative_results_cross_category}
\end{figure*}

\autoref{fig:qualitative_results_cross_category} 
provides cross-object examples. 
The other methods, which are not designed to cope with such a case, completely failed, so we omit their results. 
As the figure shows, our method successfully aligns 
(within each pair) the two different boats, the two different airplanes, and the two different chair,s even in hard settings, such as when the initialization is wrong by 180 degrees and/or when there is a significant scale difference.

\clearpage



%

\clearpage
\section{Applications}
\label{appendix:applications}

\setcounter{figure}{0}
\renewcommand{\thefigure}{B.\arabic{figure}}
We demonstrate the practical utility of our method through two downstream applications: \textit{geometrically consistent object replacement} and \textit{synchronized novel view synthesis}. These examples showcase how our method enables accurate, scale-aware alignment for real-world 3D vision tasks.

\subsection{Geometrically Consistent Object Replacement}

\begin{figure}[h!]
  \centering

  \begin{subfigure}{0.24\linewidth}
    \includegraphics[width=\linewidth]{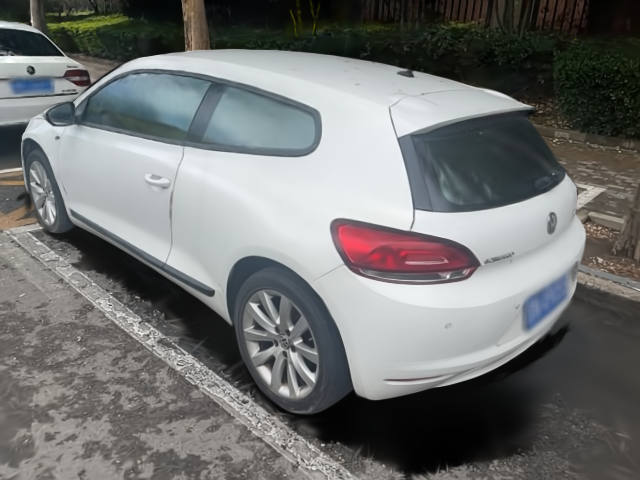}
    \caption*{White car - view 1}
  \end{subfigure}
  \hfill
  \begin{subfigure}{0.24\linewidth}
    \includegraphics[width=\linewidth]{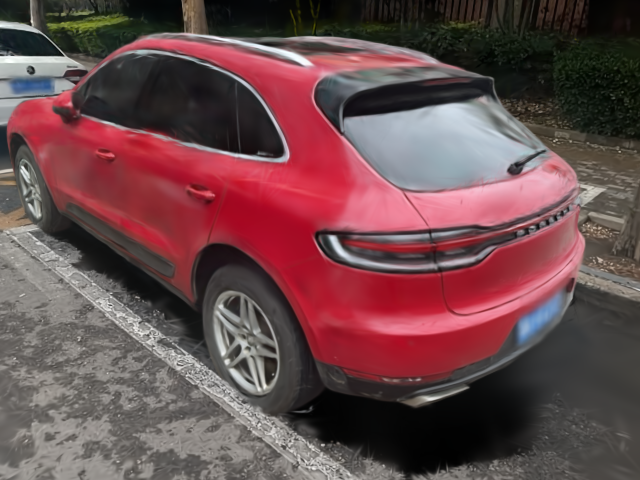}
    \caption*{Red car - view 1}
  \end{subfigure}
  \hfill
  \begin{subfigure}{0.24\linewidth}
    \includegraphics[width=\linewidth]{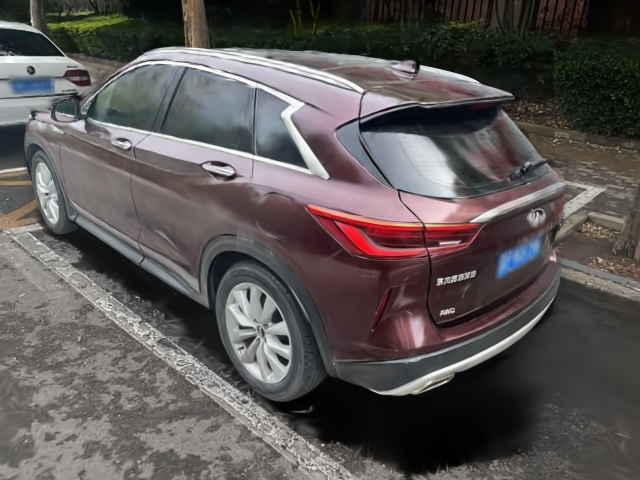}
    \caption*{Purple car - view 1}
  \end{subfigure}
  \hfill
  \begin{subfigure}{0.24\linewidth}
    \includegraphics[width=\linewidth]{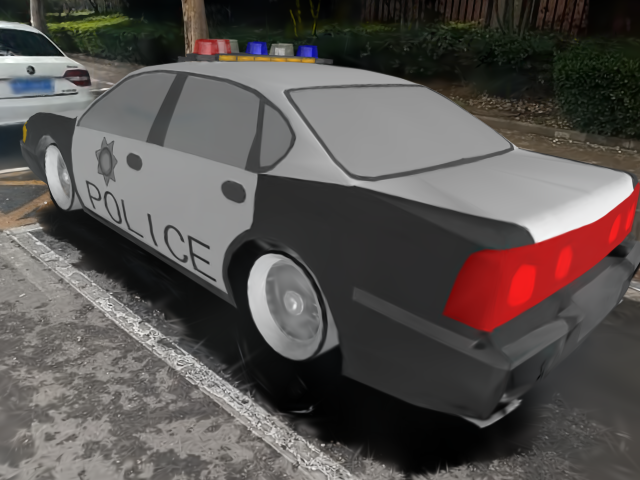}
    \caption*{Police car - view 1}
  \end{subfigure}

  \vspace{1ex} 

  \begin{subfigure}{0.24\linewidth}
    \includegraphics[width=\linewidth]{./figures/practical_app/object_replacement/00099_white.png}
    \caption*{White car - view 2}
  \end{subfigure}
  \hfill
  \begin{subfigure}{0.24\linewidth}
    \includegraphics[width=\linewidth]{./figures/practical_app/object_replacement/00099_red.png}
    \caption*{Red car - view 2}
  \end{subfigure}
  \hfill
  \begin{subfigure}{0.24\linewidth}
    \includegraphics[width=\linewidth]{./figures/practical_app/object_replacement/00099_purple.png}
    \caption*{Purple car - view 2}
  \end{subfigure}
  \hfill
  \begin{subfigure}{0.24\linewidth}
    \includegraphics[width=\linewidth]{figures/practical_app/object_replacement/00099_police.png}
    \caption*{Police car - view 2}
  \end{subfigure}

  \caption{Geometrically consistent object replacement using GSA. Each row shows a different camera view. The first column displays the original white car. The second to fourth columns show replacements with red, purple, and toy police cars, respectively. GSA aligns the models in rotation, translation, and symmetric scale, preserving their intrinsic geometry while adapting them to best match the target object's position, orientation, and scale in a structurally faithful way.}
  \label{fig:object_replacement}
\end{figure}

In this application, depicted in \autoref{fig:object_replacement}, we show the replacement of a reference object in a scene with geometrically-aligned models from the same category. Specifically, we aligned three car models (red and purple cars from 3D Real Car Dataset~\cite{du:Arxiv:2024:3DRealCar}, and a synthetic police car from ShapeNet~\cite{chang:Arxiv:2015:Shapenet}) to a reference white car from~\cite{du:Arxiv:2024:3DRealCar} using GSA. After alignment, the white car was removed from the scene and substituted with the aligned models.

Our approach ensures alignment in rotation, translation, and \emph{symmetric scale}, where the scaling factor is applied uniformly across spatial dimensions. Symmetric scaling maintains the intrinsic geometry of the source model, preserving its characteristic shape and aspect ratio while adapting its overall size to best match the target object. For example, a long and low-profile car will remain elongated after scaling, and a taller car will retain its vertical prominence. This yields visually-coherent substitutions that are structurally faithful.

\clearpage

\subsection{Synchronized Novel View Synthesis}

We demonstrate synchronized novel view synthesis across semantically similar objects by aligning all models to a reference using GSA (e.g., aligning all the objects to object 1). This alignment enables consistent view generation from shared camera poses using 3DGS.

By resolving differences in orientation and scale, our method produces spatially coherent renderings beneficial for applications such as data visualization, robotic perception, and more.

\begin{figure}[h]
    \centering

    \hspace{1.4cm}
    \begin{minipage}{0.195\linewidth}\centering \textbf{Object 1} \end{minipage}%
    \hspace{1pt}
    \begin{minipage}{0.195\linewidth}\centering \textbf{Object 2} \end{minipage}%
    \hspace{1pt}
    \begin{minipage}{0.195\linewidth}\centering \textbf{Object 3} \end{minipage}%
    \hspace{1pt}
    \begin{minipage}{0.195\linewidth}\centering \textbf{Object 4} \end{minipage}

    \vspace{-10pt}

    \noindent
    \begin{minipage}[c][3.2cm][c]{1.4cm}
        \centering \textbf{Novel\\View 1}
    \end{minipage}%
    \begin{minipage}[c][3.2cm][c]{0.195\linewidth}
        \includegraphics[width=\linewidth]{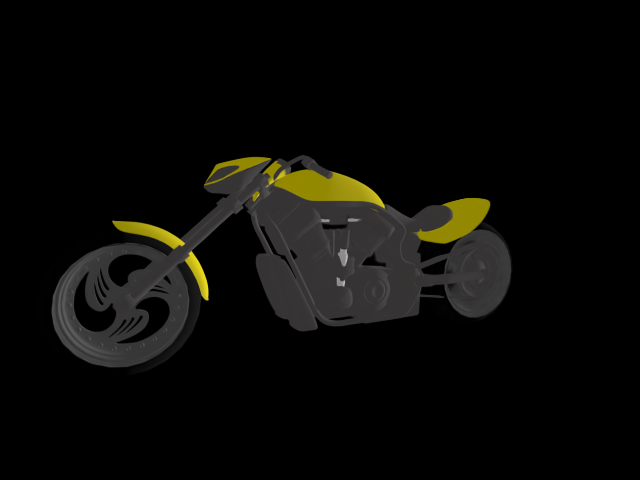}
    \end{minipage}%
    \hspace{1pt}
    \begin{minipage}[c][3.2cm][c]{0.195\linewidth}
        \includegraphics[width=\linewidth]{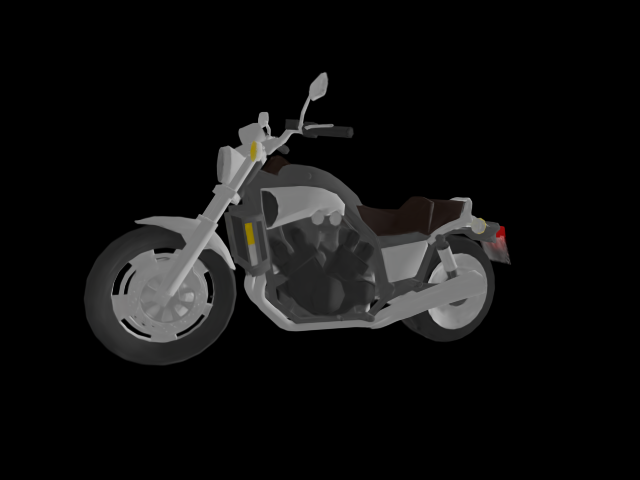}
    \end{minipage}%
    \hspace{1pt}
    \begin{minipage}[c][3.2cm][c]{0.195\linewidth}
        \includegraphics[width=\linewidth]{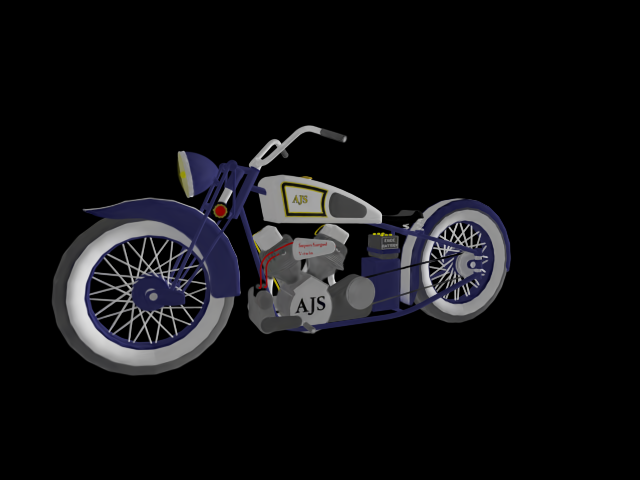}
    \end{minipage}%
    \hspace{1pt}
    \begin{minipage}[c][3.2cm][c]{0.195\linewidth}
        \includegraphics[width=\linewidth]{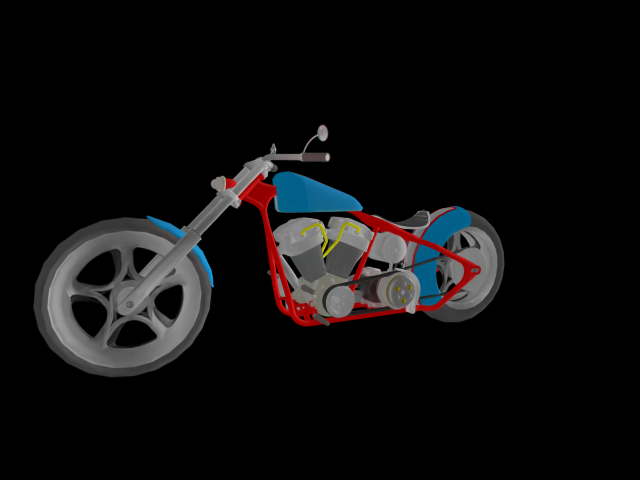}
    \end{minipage}

    \vspace{-30pt}

    \noindent
    \begin{minipage}[c][3.2cm][c]{1.4cm}
        \centering \textbf{Novel\\Feature\\View 1}
    \end{minipage}%
    \begin{minipage}[c][3.2cm][c]{0.195\linewidth}
        \includegraphics[width=\linewidth]{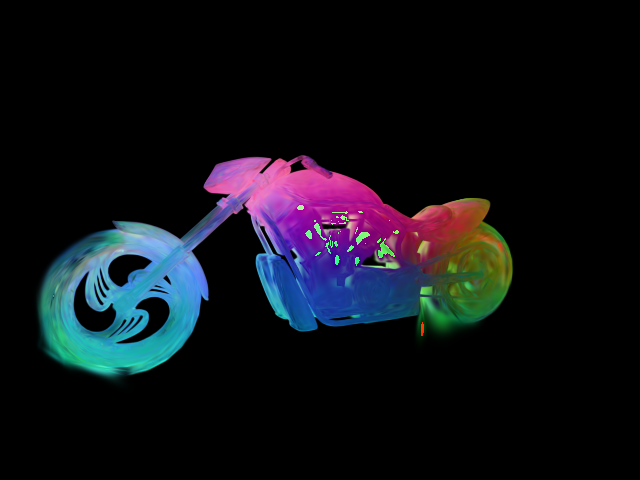}
    \end{minipage}%
    \hspace{1pt}
    \begin{minipage}[c][3.2cm][c]{0.195\linewidth}
        \includegraphics[width=\linewidth]{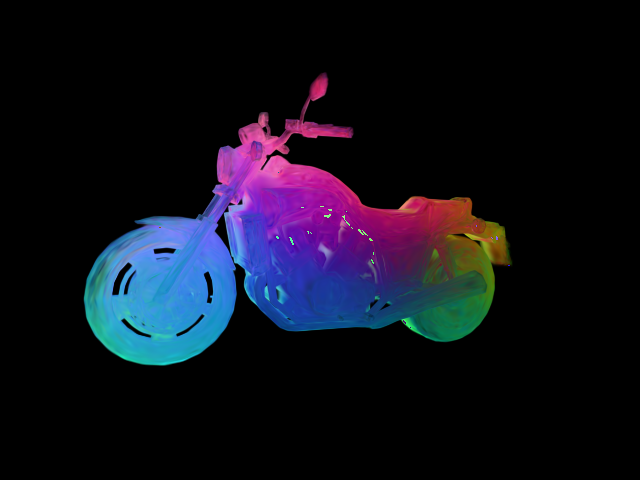}
    \end{minipage}%
    \hspace{1pt}
    \begin{minipage}[c][3.2cm][c]{0.195\linewidth}
        \includegraphics[width=\linewidth]{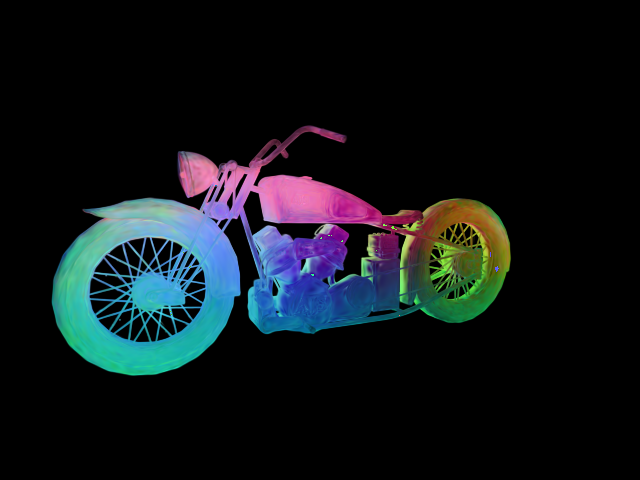}
    \end{minipage}%
    \hspace{1pt}
    \begin{minipage}[c][3.2cm][c]{0.195\linewidth}
        \includegraphics[width=\linewidth]{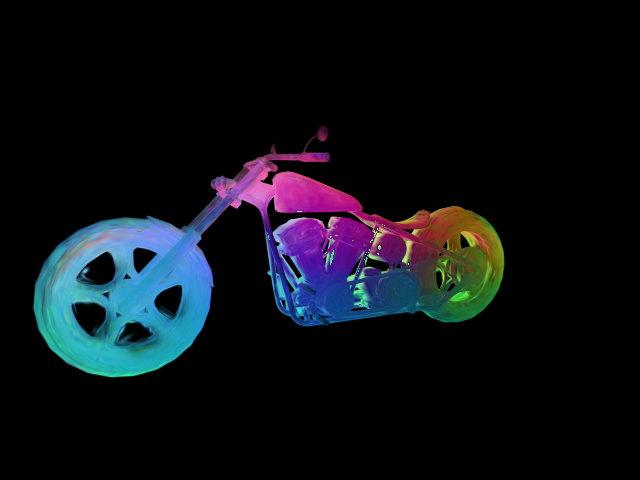}
    \end{minipage}

    \vspace{-30pt}

    \noindent
    \begin{minipage}[c][3.2cm][c]{1.4cm}
         \centering \textbf{Novel\\View 2}
    \end{minipage}%
    \begin{minipage}[c][3.2cm][c]{0.195\linewidth}
        \includegraphics[width=\linewidth]{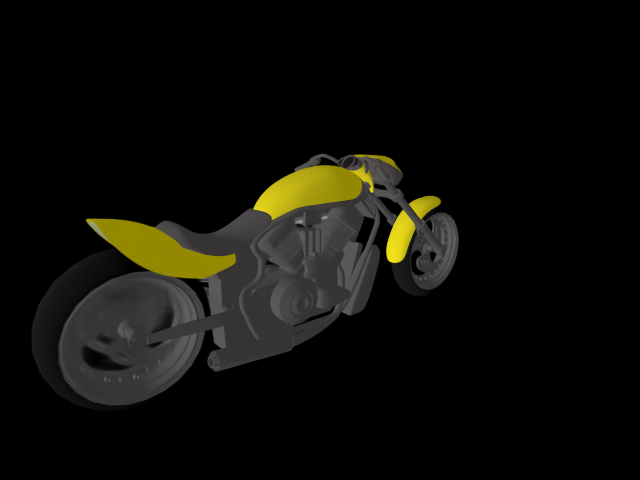}
    \end{minipage}%
    \hspace{1pt}
    \begin{minipage}[c][3.2cm][c]{0.195\linewidth}
        \includegraphics[width=\linewidth]{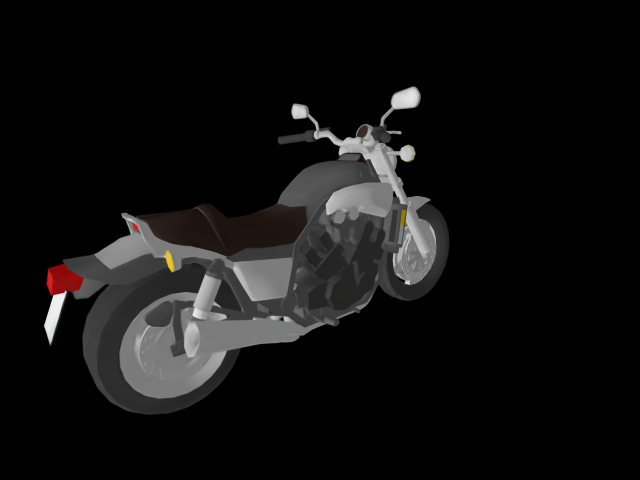}
    \end{minipage}%
    \hspace{1pt}
    \begin{minipage}[c][3.2cm][c]{0.195\linewidth}
        \includegraphics[width=\linewidth]{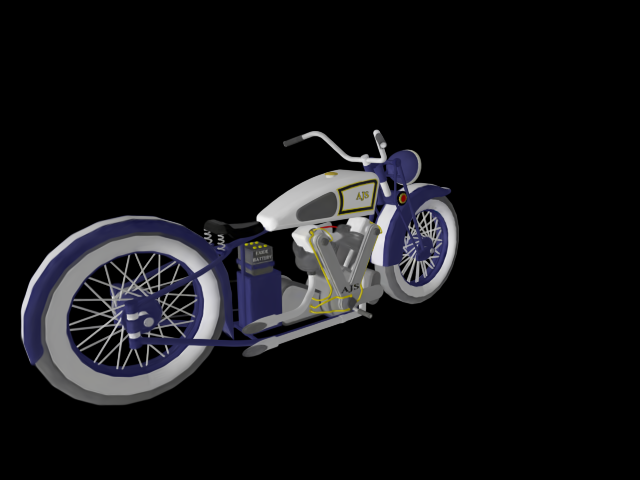}
    \end{minipage}%
    \hspace{1pt}
    \begin{minipage}[c][3.2cm][c]{0.195\linewidth}
        \includegraphics[width=\linewidth]{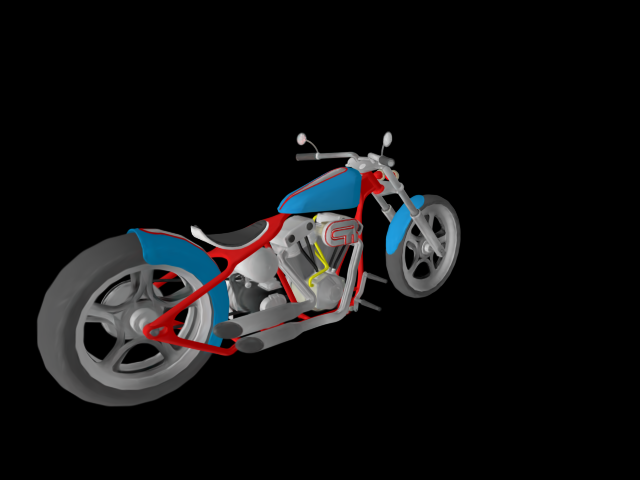}
    \end{minipage}

    \vspace{-30pt}

    \noindent
    \begin{minipage}[c][3.2cm][c]{1.4cm}
        \centering \textbf{Novel\\Feature\\View 2}
    \end{minipage}%
    \begin{minipage}[c][3.2cm][c]{0.195\linewidth}
        \includegraphics[width=\linewidth]{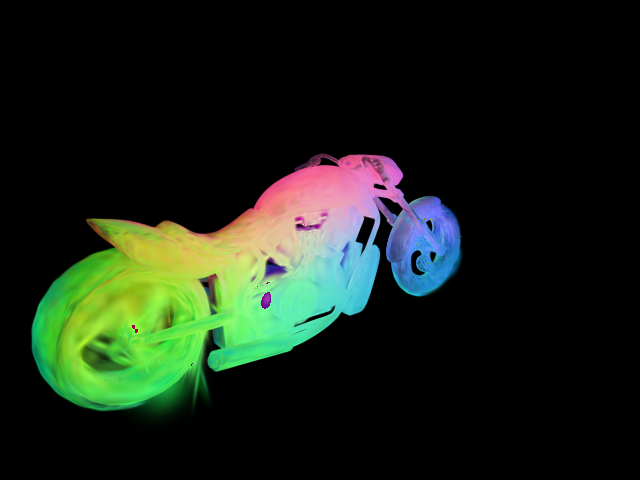}
    \end{minipage}%
    \hspace{1pt}
    \begin{minipage}[c][3.2cm][c]{0.195\linewidth}
        \includegraphics[width=\linewidth]{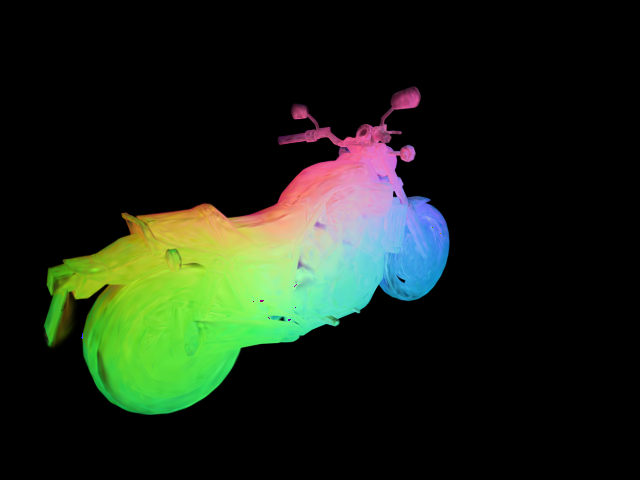}
    \end{minipage}%
    \hspace{1pt}
    \begin{minipage}[c][3.2cm][c]{0.195\linewidth}
        \includegraphics[width=\linewidth]{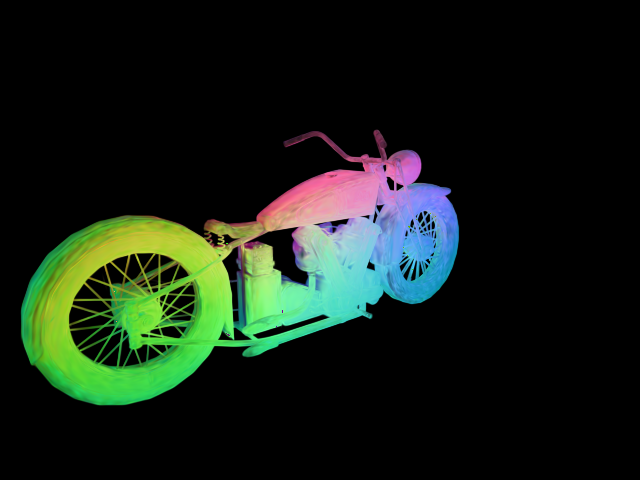}
    \end{minipage}%
    \hspace{1pt}
    \begin{minipage}[c][3.2cm][c]{0.195\linewidth}
        \includegraphics[width=\linewidth]{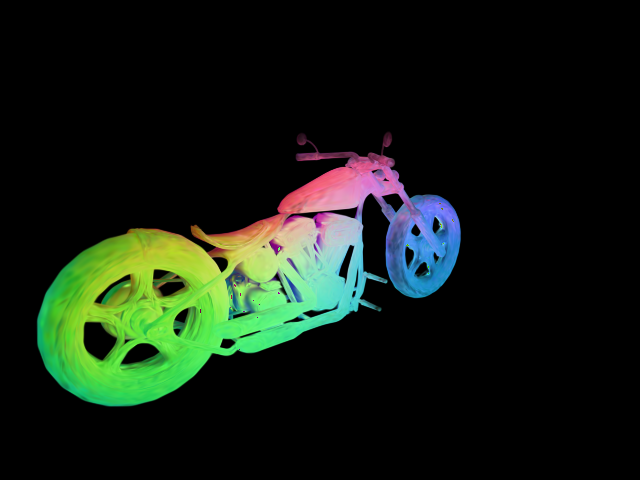}
    \end{minipage}

    \vspace{-15pt}

    \caption{Synchronized novel view and feature renderings across multiple semantically aligned motorcycle instances. Each row shows either a novel rendering or the corresponding feature projection for different objects and camera views.}
    \label{fig:sync_novelview}
\end{figure}

\begin{figure}[h]
    \centering
    \begin{minipage}[c]{0.44\linewidth}
        \centering
        \includegraphics[width=\linewidth]{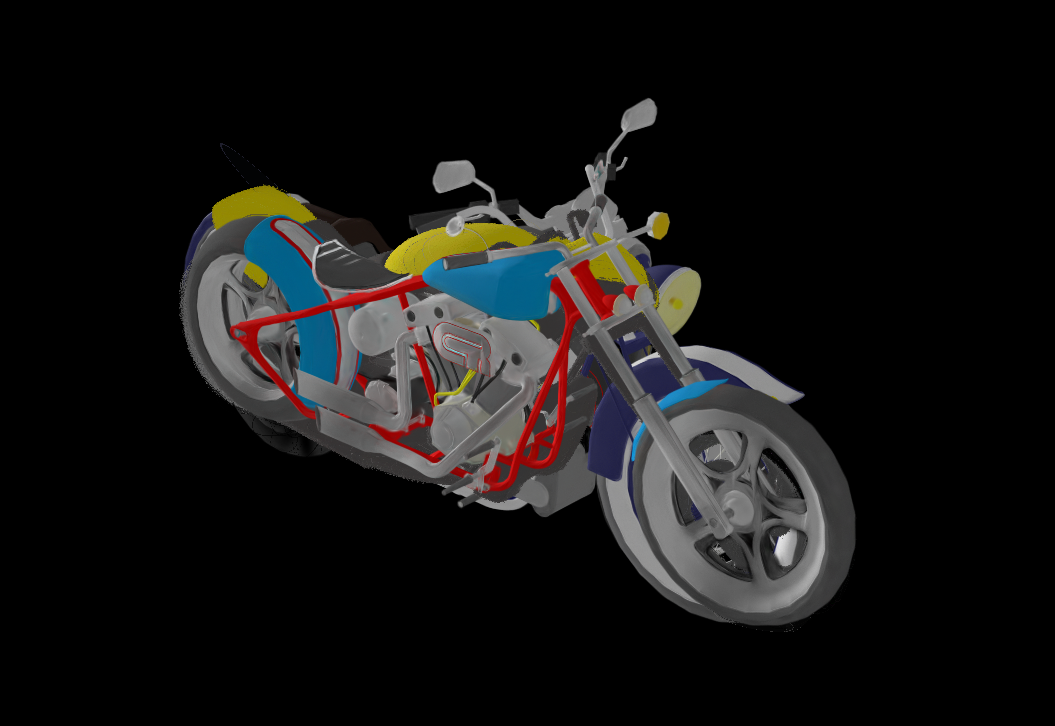}
        \smallskip
        \textbf{Combined Aligned Models - Novel View 1}
    \end{minipage}%
    \hspace{0.04\textwidth}
    \begin{minipage}[c]{0.44\linewidth}
        \centering
        \includegraphics[width=\linewidth]{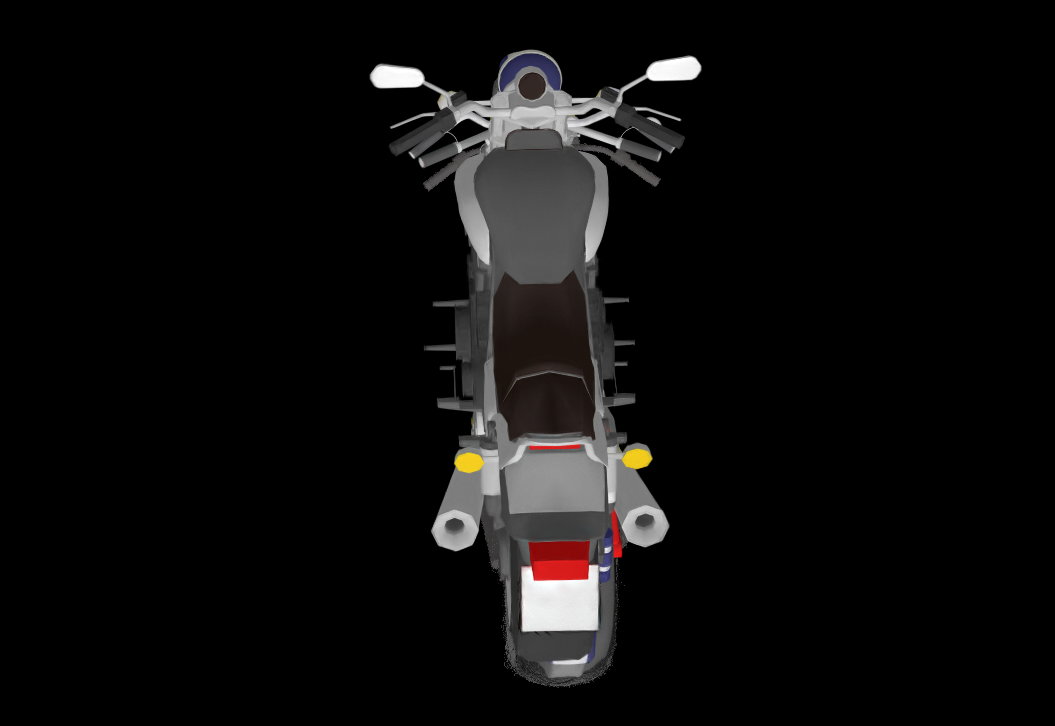}
        \smallskip
        \textbf{Combined Aligned Models - Novel View 2}
    \end{minipage}

    \caption{Renderings of the unified 3D Gaussian Splatting (3DGS) model combining all gaussians from the aligned motorcycle instances, shown from two additional novel synchronized views. These renderings provide a global perspective of the unified category-level geometry.}
    \label{fig:combined_sync_novelview}
\end{figure}

\autoref{fig:sync_novelview} shows synchronized views across instances within the same category. \autoref{fig:combined_sync_novelview} visualizes the jointly-aligned models from two additional novel views.

\clearpage
\section{Scale Estimation}
\label{appendix:scale_estimation}

\setcounter{figure}{0}
\renewcommand{\thefigure}{C.\arabic{figure}}

\renewcommand{\thetable}{C.\arabic{table}}
\setcounter{table}{0}
In \autoref{appendix:additional_visual_results}, we presented qualitative scale estimation results across objects within the same category. In such cases, scale and translation are inherently ambiguous due to intra-category variation in size and geometry, leading to multiple plausible alignment solutions and a one-to-many mapping in the transformation space.

To quantitatively assess our method under well-defined conditions, we trained 30 3D Gaussian Splatting (3DGS)~\cite{kerbl:ACM:2023:3DGS}  models on distinct ShapeNet ~\cite{chang:Arxiv:2015:Shapenet} objects. Each object was duplicated and randomly transformed using translation, rotation (up to 180° per axis), and scaling (up to ×10). We then registered each transformed model to its original to evaluate scale estimation accuracy.

\paragraph{Scale Error Metric.}
We report the \textit{relative scale error}, computed as the absolute difference between the estimated scale $s_{\text{est}}$ and the ground-truth scale $s_{\text{gt}}$, normalized by the ground-truth value and expressed as a percentage:
\[
\text{Scale Error} = \frac{|s_{\text{est}} - s_{\text{gt}}|}{s_{\text{gt}}} \times 100\%.
\]
This metric penalizes both under- and over-estimation symmetrically and provides an interpretable measure of scale estimation quality.

\begin{table}[h]
\centering
\caption{Quantitative evaluation of rotation and scale estimation on transformed ShapeNet objects.}
\label{tab:registration_results}
\begin{tabular}{lcc}
\toprule
\textbf{Method} & \textbf{Mean RRE (deg)} & \textbf{Mean Scale Error (\%)} \\
\midrule
GSA: Coarse & $0.362 $ & $0.21$ \\
GSA: Coarse+Fine & $0.074$ & $0.08$ \\
\bottomrule
\end{tabular}
\end{table}
As shown in \autoref{tab:registration_results}, our method achieves highly accurate rotation and scale estimation. The coarse stage yields a scale error of 0.21\% and a rotation error of 0.362°. With fine alignment, these improve to 0.08\% and 0.074°, respectively.

\begin{figure}[h]
    \centering

    \newcommand{\topW}{0.3\linewidth}
    \newcommand{\botW}{0.27\linewidth}

    \makebox[\textwidth][c]{%
        \begin{minipage}[t]{\topW}
            \centering
            \textbf{Random Initialization}\\[3pt]
            \includegraphics[width=\linewidth]{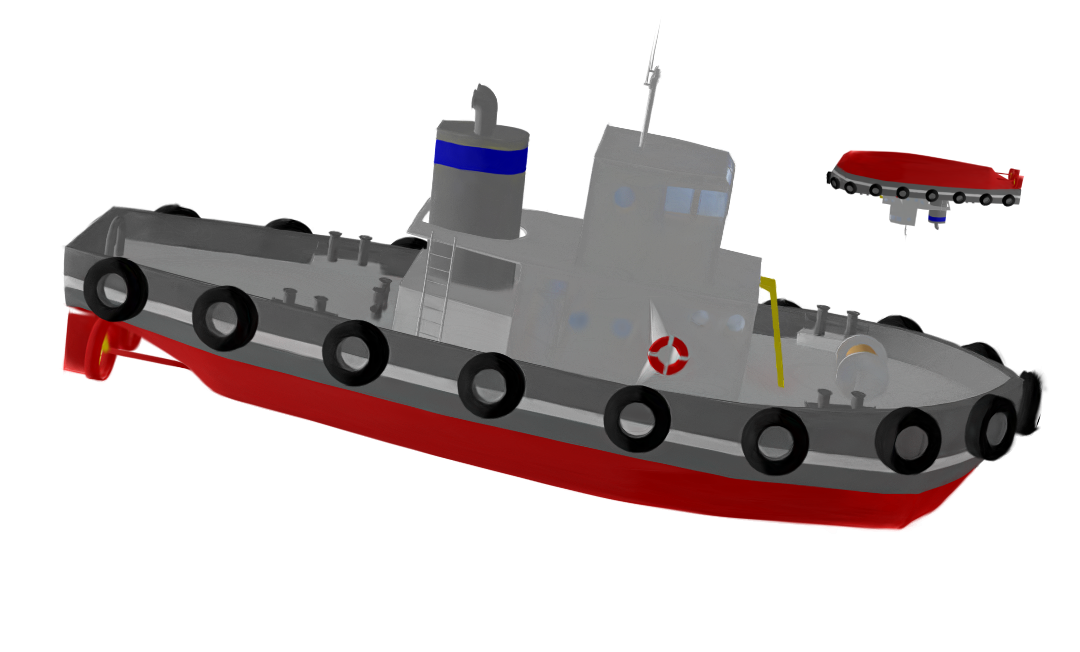}
        \end{minipage}
        \hspace{4pt}
        \begin{minipage}[t]{\topW}  
            \centering
            \textbf{Coarse Result}\\[3pt]
            \includegraphics[width=\linewidth]{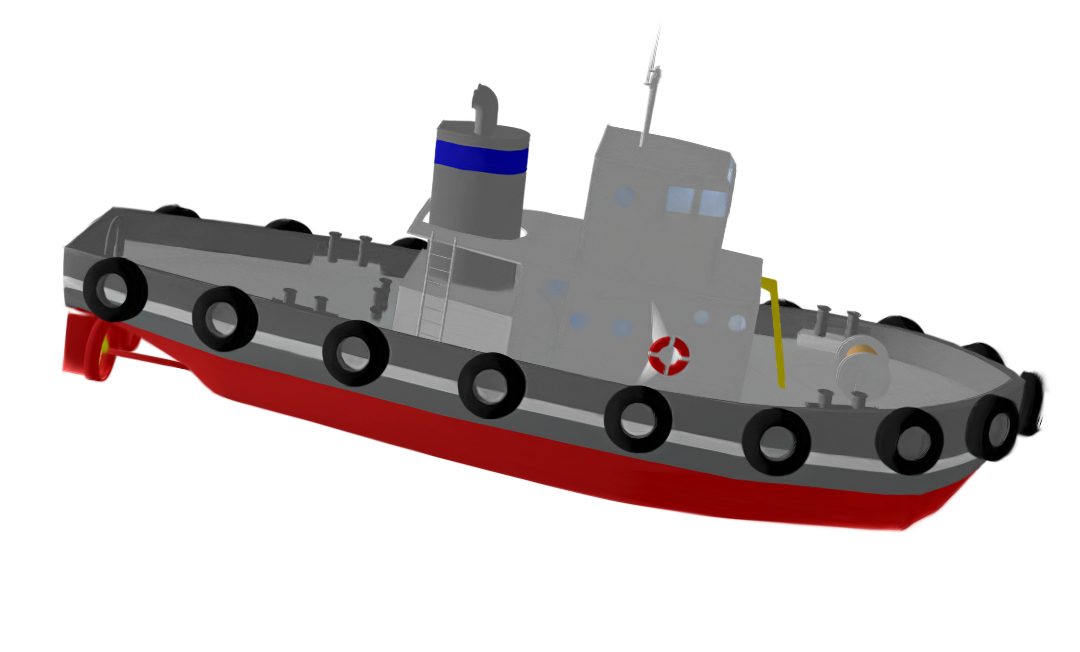}
        \end{minipage}
        \hspace{4pt}
        \begin{minipage}[t]{\topW}
            \centering
            \textbf{Fine Result}\\[3pt]
            \includegraphics[width=\linewidth]{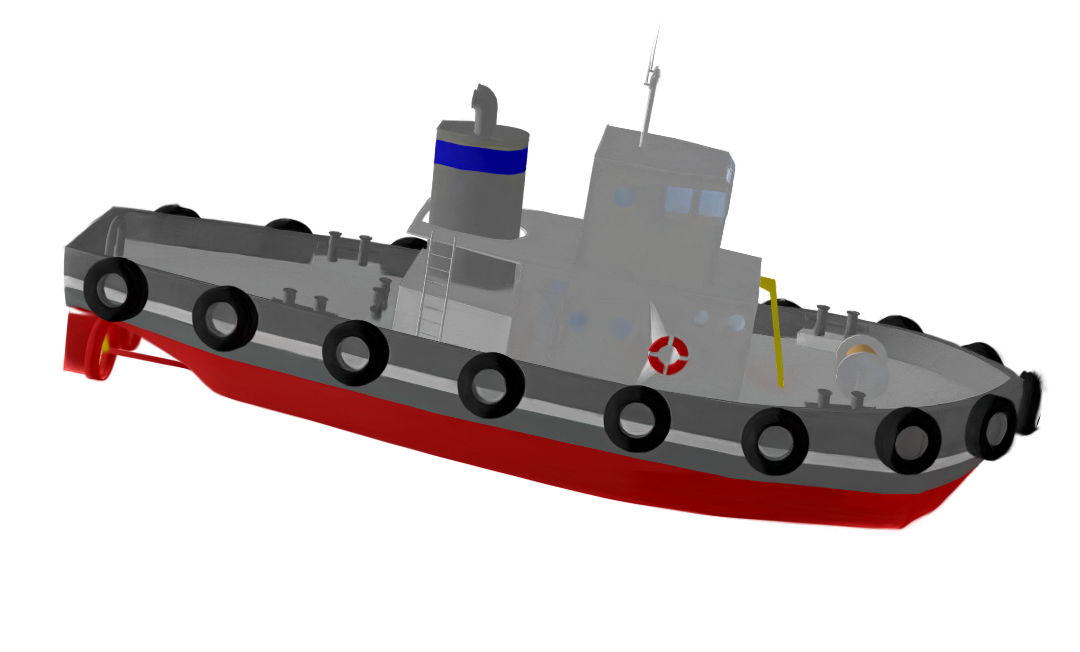}
        \end{minipage}
    }

    \vspace{0.7em}

    \makebox[\textwidth][c]{%
        \begin{minipage}[t]{\botW}
            \centering
            \textbf{Zoom Coarse}\\[3pt]
            \includegraphics[width=\linewidth]{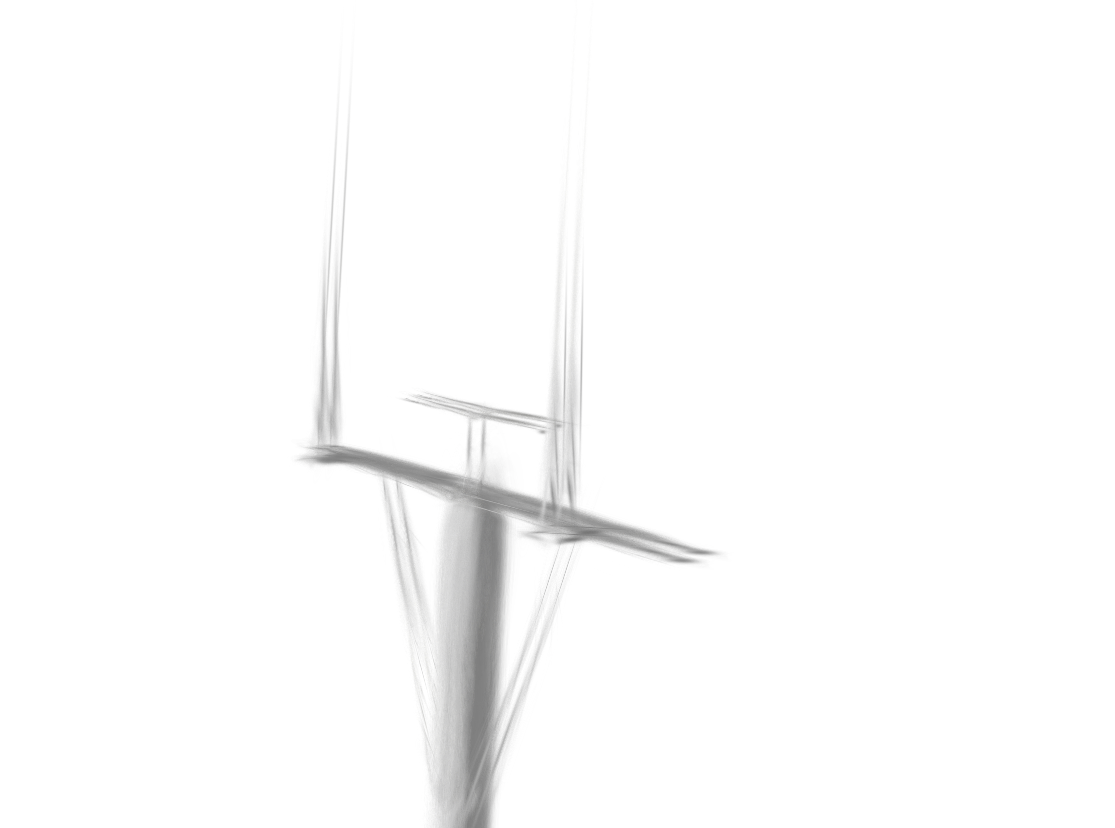}
        \end{minipage}
        \hspace{8pt}
        \begin{minipage}[t]{\botW}
            \centering
            \textbf{Zoom Fine}\\[3pt]
            \includegraphics[width=\linewidth]{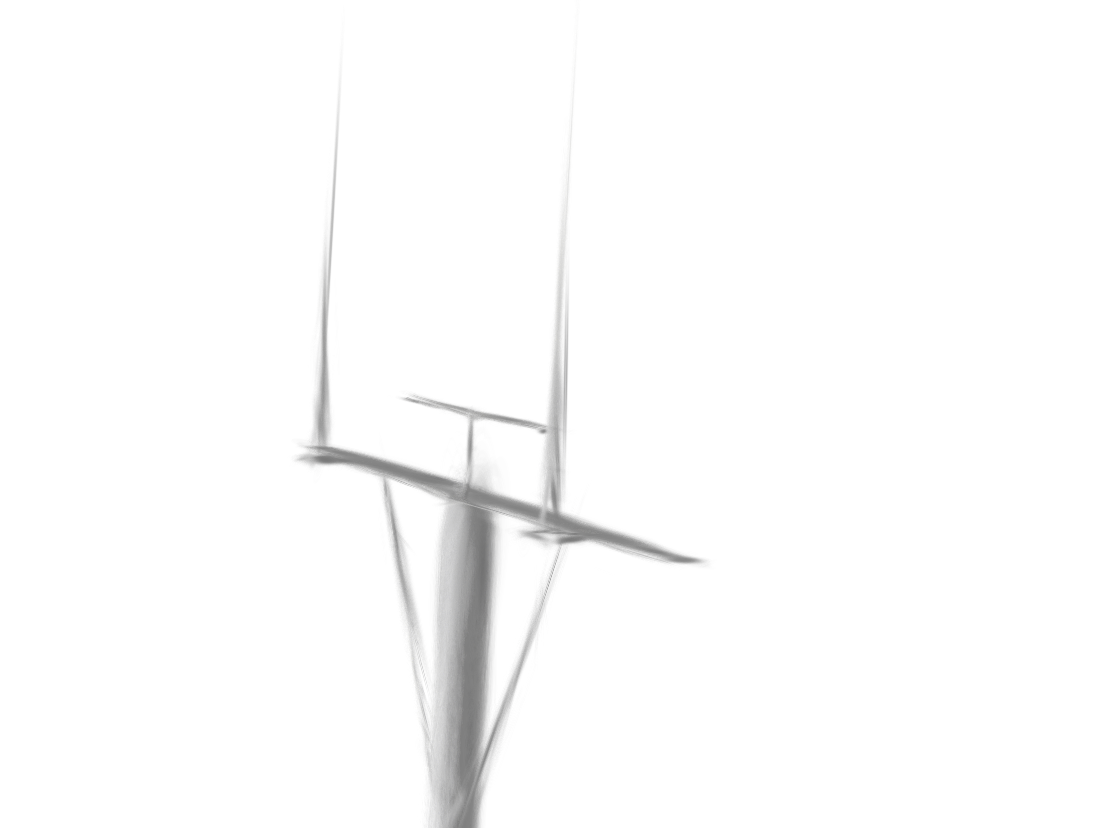}
        \end{minipage}
    }

    \caption{Qualitative registration results. Top row: comparison of the random initialization, coarse, and fine alignment results. Bottom row: close-up views of the boat top, highlighting the improved accuracy of the fine stage over the coarse alignment.}
    \label{fig:scale_registration_boat_qualitative}
\end{figure}

To qualitatively support these results, we show in \autoref{fig:scale_registration_boat_qualitative} an example with a large initial scale discrepancy and nearly 180° misalignment along two axes. While the coarse alignment already yields high accuracy, close-up views of the boat top reveal subtle yet consistent improvements achieved by the fine stage, further validating the quantitative gains in both scale and rotation.

\clearpage
\section{The Impact the Choice of Features has on the Alignment }
\label{appendix:features_impact}

\setcounter{figure}{0}
\renewcommand{\thefigure}{D.\arabic{figure}}

We now further explain 
our choice of using
the features from~\cite{Mariotti:CVPR:2024:ViewpointSphereMap} and not, say, other plausible choices
such as DINOv2~\cite{Oquab:arXiv:2023:DINOv2}
or TellingLeftfromRight~\cite{Zhang:CVPR:2024:TellingLeftRight}. 

\begin{figure}[h]
\centering

\newcommand{\fixedwidthcite}[1]{\raisebox{8mm}{\makebox[8mm][c]{#1}}}

\fixedwidthcite{\cite{Oquab:arXiv:2023:DINOv2}} 
\includegraphics[trim=0 22mm 0 20mm, clip, width=0.45\linewidth]{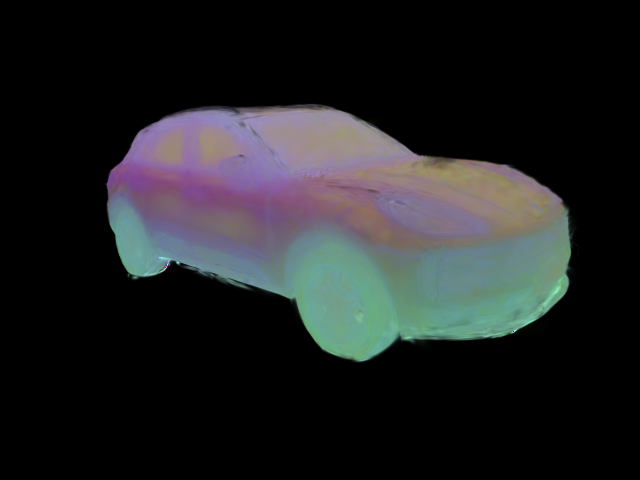} \hspace{-3mm}
\includegraphics[trim=0 22mm 0 20mm, clip, width=0.45\linewidth]{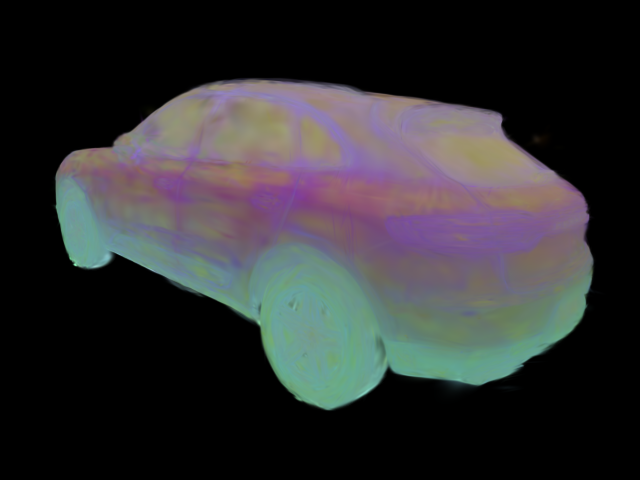} \\[1mm]

\fixedwidthcite{\cite{Zhang:CVPR:2024:TellingLeftRight}}
\includegraphics[trim=0 22mm 0 20mm, clip, width=0.45\linewidth]{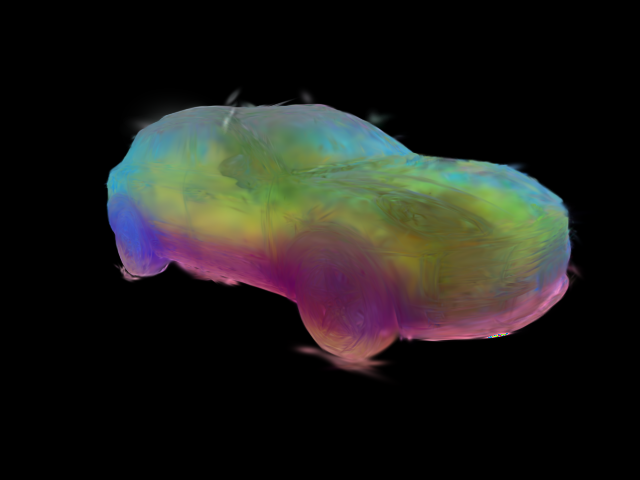} \hspace{-3mm}
\includegraphics[trim=0 22mm 0 20mm, clip, width=0.45\linewidth]{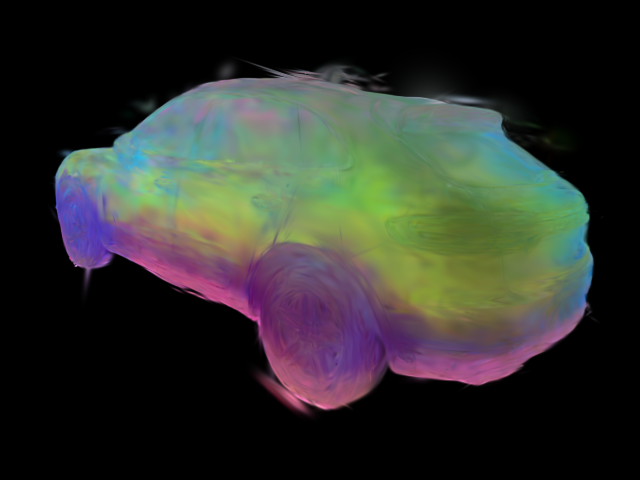} \\[1mm]  

\fixedwidthcite{\cite{Mariotti:CVPR:2024:ViewpointSphereMap}}
\includegraphics[trim=0 22mm 0 20mm, clip, width=0.45\linewidth]{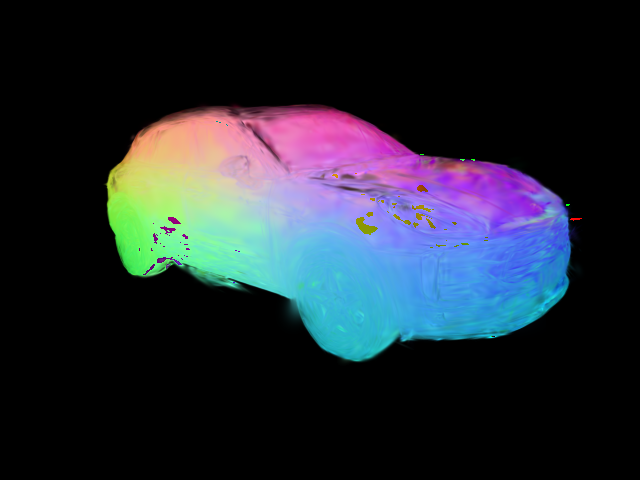} \hspace{-3mm}
\includegraphics[trim=0 22mm 0 20mm, clip, width=0.45\linewidth]{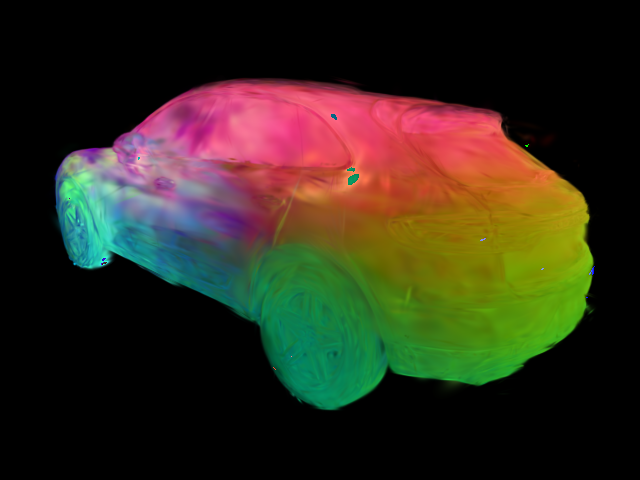} \\[2mm]  

\makebox[0.45\linewidth]{View 1, Rendered} \hspace{-3mm}
\makebox[0.45\linewidth]{View 2, Rendered} \\

\caption{Augmented 3DGS models 
using different choices of semantic features:
\cite{Mariotti:CVPR:2024:ViewpointSphereMap}, \cite{Oquab:arXiv:2023:DINOv2}, and \cite{Zhang:CVPR:2024:TellingLeftRight}. 
For~\cite{Oquab:arXiv:2023:DINOv2} or 
\cite{Zhang:CVPR:2024:TellingLeftRight},  dimensionality 
was reduced to 3 by Principal Component Analysis.}
\label{fig:ablation_enhanced_features}
\end{figure}

We consider three feature types: DINOv2~\cite{Oquab:arXiv:2023:DINOv2}, 
which is not geometry-aware, 
and the geometric-aware
TellingLeftfromRight~\cite{Zhang:CVPR:2024:TellingLeftRight} and view-guided spherical map features~\cite{Mariotti:CVPR:2024:ViewpointSphereMap}. 
\autoref{fig:ablation_enhanced_features} shows that DINOv2 
struggles with distinguishing between semantically similar but spatially distinct parts (\eg, all four wheels and the hood/trunk). This is issue for alignment as it leads to ambiguity and up to \emph{180-degree misalignment}.
Similarly, TellingLeftfromRight, while effective for some categories, for others degenerates to image-level left–right prediction instead of object-centric geometric reasoning, and therefore does not resolve \emph{cross-image symmetry ambiguities}, which can lead to 180-degree errors.
In other words, while DINOv2 and TellingLeftfromRight operate in higher-dimensional spaces (\eg, 384 and 768 for ViTs and ViT-Base, resp.) and encode rich semantics, their utility for alignment is limited. 
In contrast, \emph{view-guided spherical map features} provide effective geometry-aware representations, leading to significantly improved alignment. Moreover, these features are much more compact (3D feature vectors), thereby ensuring 
not only geometric awareness but also efficiency.

\FloatBarrier
\newpage

\begin{figure}[h]
\centering

\includegraphics[width=0.48\linewidth]{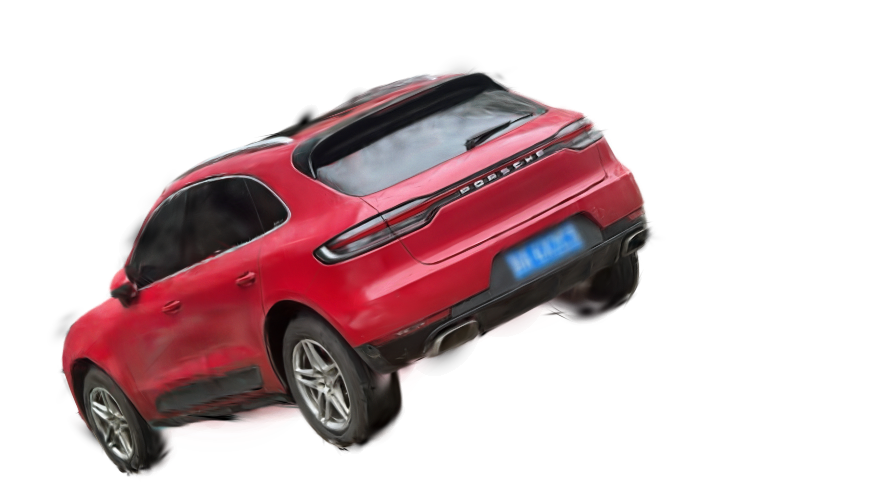} \hspace{1mm}
\includegraphics[width=0.48\linewidth]{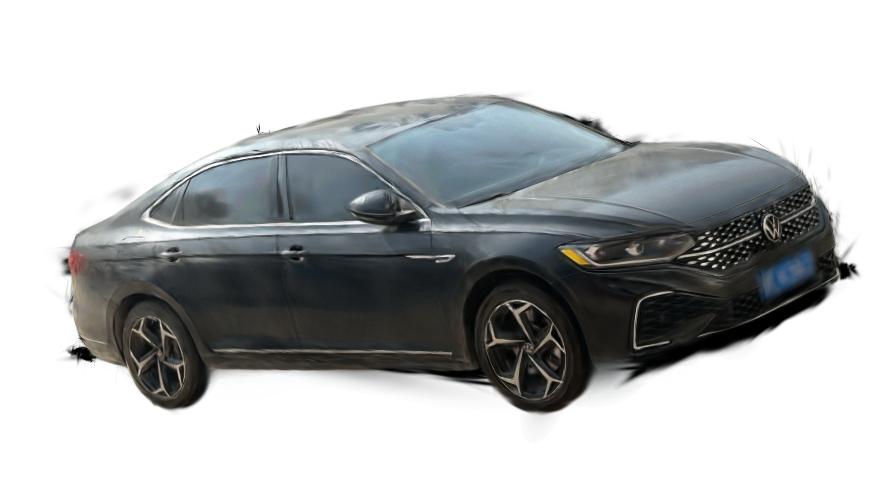} \\[3mm]

\includegraphics[width=0.48\linewidth]{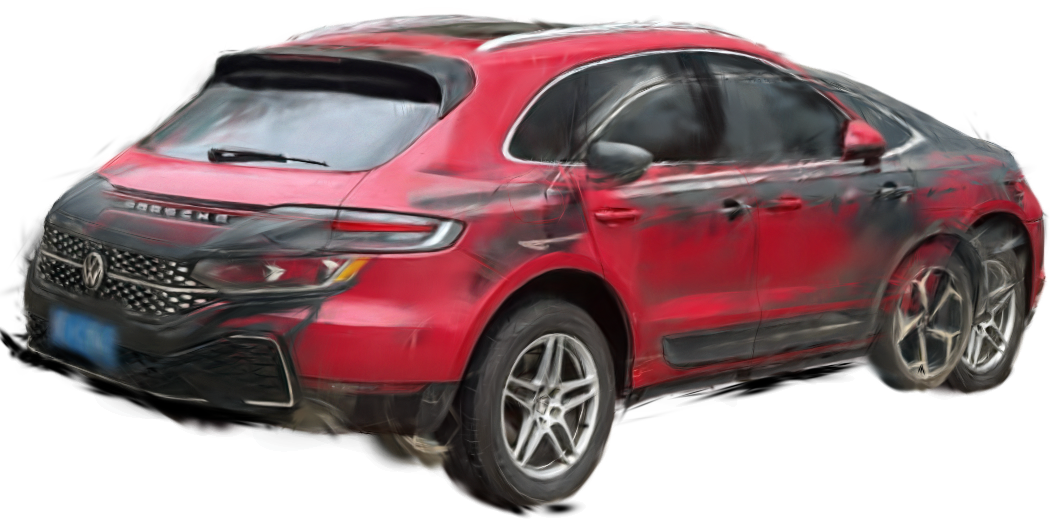}
\hspace{1mm}
\includegraphics[width=0.48\linewidth]{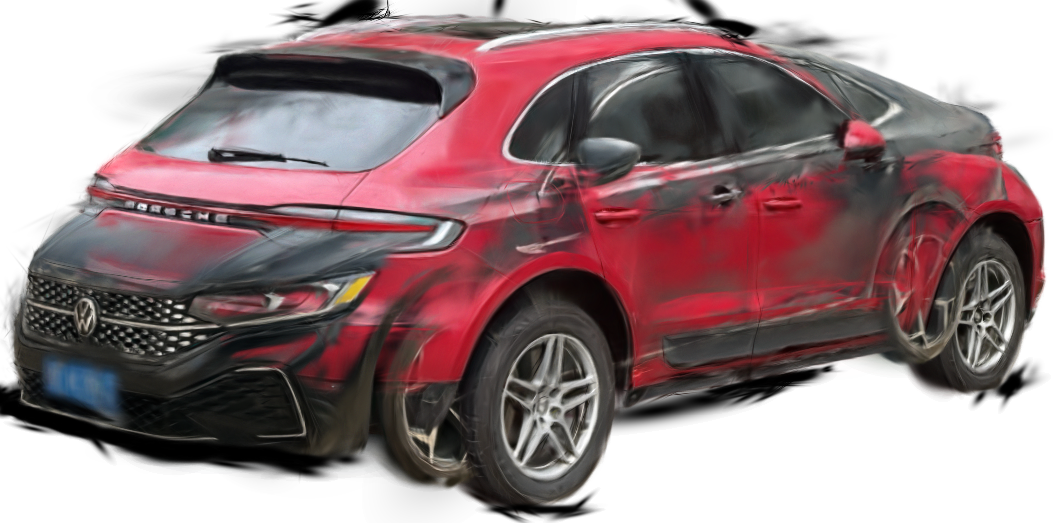} \\
{\small \cite{Oquab:arXiv:2023:DINOv2}} 
\hspace{40mm} 
{\small \cite{Zhang:CVPR:2024:TellingLeftRight}}  \\[3mm]

\includegraphics[width=0.6\linewidth]{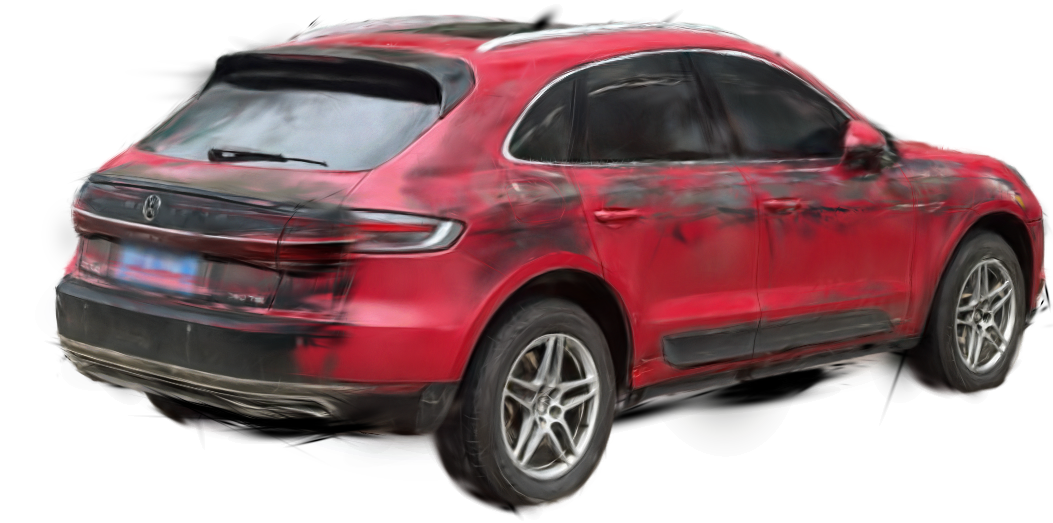} \\
\cite{Mariotti:CVPR:2024:ViewpointSphereMap} \\[3mm]

\caption{Qualitative comparison 
of the GSA coarse alignment step, when using different features. 
Top row: Unaligned models. 
Middle row, left: Alignment result
when using DINOv2~\cite{Oquab:arXiv:2023:DINOv2}. 
Note the angular mistake is about 180 degrees. 
Middle row, right: Alignment 
results when using 
and \cite{Zhang:CVPR:2024:TellingLeftRight}.
Again the mistake is about 180 degrees, 
and the tilting angle is also evidently wrong. 
Bottom row: In contrast, when GSA uses~\cite{Mariotti:CVPR:2024:ViewpointSphereMap}, it successfully aligns the cars.}
\label{Fig:CompareFeatureEffectOnAlign}

\end{figure}

A typical comparison between
these three feature types,
when used within the proposed GSA method, 
is shown in 
\autoref{Fig:CompareFeatureEffectOnAlign}. 
As explained in the figure's caption,
in this example,
using the features
from either~\cite{Oquab:arXiv:2023:DINOv2}
or~\cite{Zhang:CVPR:2024:TellingLeftRight} results in gross errors, with angular errors close to 180 degrees.
This is because these features
do not disambiguate or do not disambiguate
enough to distinguish the car directions (e.g., left, right, front, back).
When we leverage the features from~\cite{Mariotti:CVPR:2024:ViewpointSphereMap}, however, the problem disappears. 

\clearpage
\section{The Common problem in 3DGS Models of 
      Background-related Gaussians
      and our Effective solution to it}
\label{appendix:black_gaussians}

\setcounter{figure}{0}
\renewcommand{\thefigure}{E.\arabic{figure}}

\begin{figure}[h]
\centering

\newcommand{\imgW}{0.8\linewidth}   

\subcaptionbox{3DGS with spurious background Gaussians}{%
    \includegraphics[width=\imgW]{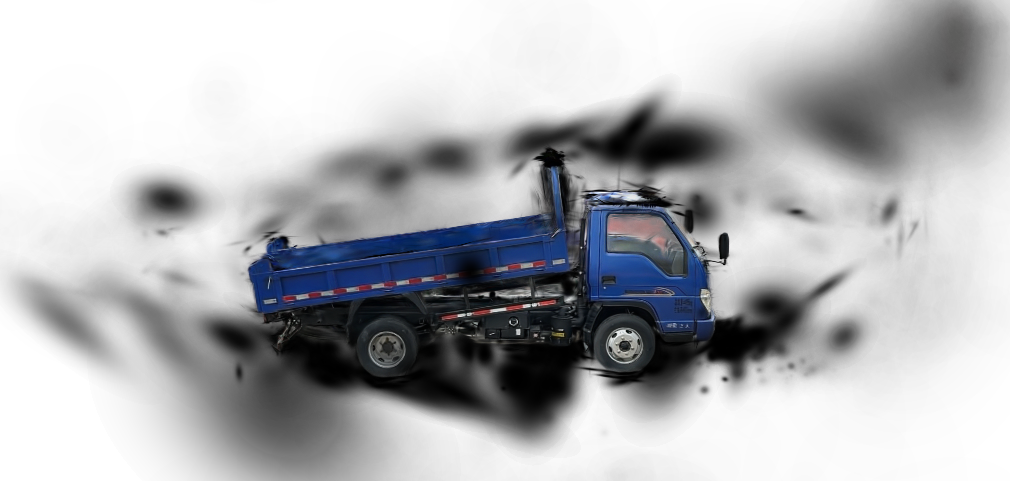}
}

\subcaptionbox{3DGS when our solution for eliminating background Gaussian is used.}{%
    \includegraphics[width=\imgW]{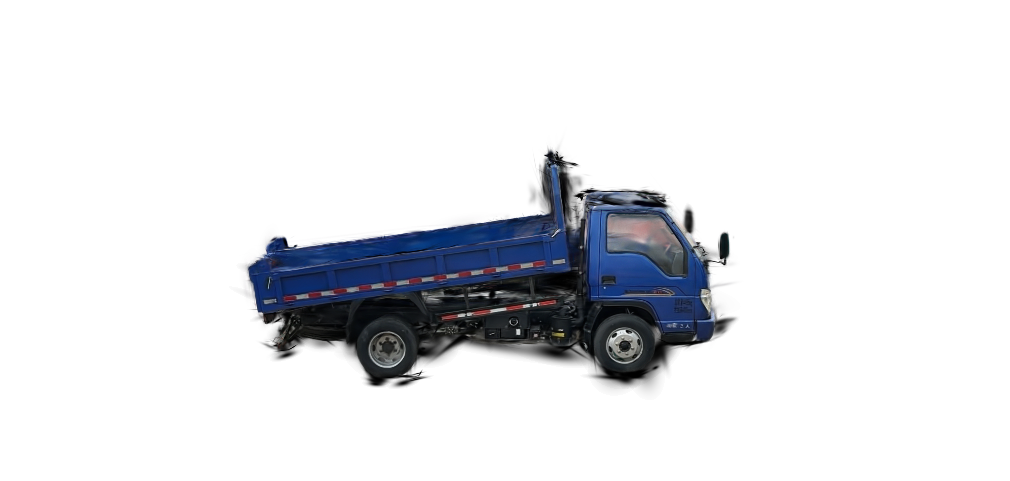}
}

\caption{The effect of background pruning. (a) Extraneous background Gaussians create occlusions, and are inconsistent across the two models, complicating alignment. (b) Our solution removes these artifacts, improving alignment consistency.}
\label{fig:black_gaussians_ablation}
\end{figure}

\autoref{fig:black_gaussians_ablation}
 shows that this phenomenon 
can hurt results if left unaddressed. For example the black Gaussians can block novel views of the object. 
While this issue affects 3DGS models in general, it is particularly problematic in our case because such occlusions directly degrade the multi-view consistency loss during fine alignment.

As a solution,  which 
is in fact useful not just for our case but in general when working with 3DGS models, 
we apply a simple effective strategy. 
When building the 3DGS model,
in each densification 
step (a standard step in 3DGS construction),
we remove the Gaussians that do not encode meaningful features. Specifically, if a Gaussian has a near-zero feature vector
(as measured by its magnitude), we remove it. This approach eliminates unwanted artifacts, improves robustness, and enhances the overall 3D reconstruction quality.

\clearpage
\section{Computational Complexity and Runtime Analysis}
\label{appendix:complexity_runtime}

Our current implementation of coarse alignment is fully GPU-accelerated, achieving runtimes of less than one second. The subsequent fine registration step adds approximately 6--8 seconds. During the coarse stage, we compute a subset of potential correspondences for each point in model~1 based on semantic feature similarity, with a complexity of \(O(NM)\), where \(N\) and \(M\) are the number of Gaussians in the first and second models, respectively. We then perform iterative nearest neighbor searches within these subsets, constrained to points with at least one valid match. This search runs in \(O(SK_{\max})\), where \(S\) is the number of such points and \(K_{\max}\) is the maximum size of any correspondence subset.

While this GPU implementation provides excellent performance, we plan to contribute an optimized multithreaded CPU version of the coarse alignment algorithm to the Open3D~\cite{zhou:Arxiv:2018:open3d} framework. This will broaden accessibility and ensure high-performance semantic feature-based registration even in CPU-only environments.

\clearpage
\section{Hybrid Kabsch-Umeyama and Horn Closed-Form Solution to the Absolute Orientation Problem}
\label{appendix:closed_form_solutions}

The following describes a closed-form solution combining the Kabsch-Umeyama method~\cite{umeyama1991least}
for optimal rotation and translation with Horn's method for estimating a symmetric scale~\cite{horn1987closedquaternion,horn1988closed}.
Let \( \bX = \{ \bx_i \}_{i=1}^{N} \) and \( \bY = \{ \by_i \}_{i=1}^{N} \) be two sets of corresponding points in \(\mathbb{R}^3\). The goal is to find a similarity transformation (rotation \( \bR \), scale \( s \), and translation \( \bt \)) aligning \( \bX \) to \( \bY \).

\subsection{Problem Formulation}

The optimal similarity transformation minimizes:
\begin{align}
    E(s, \bR, \bt) = \sum_{i=1}^{N} \| s\bR \bx_i + \bt - \by_i \|^2.
\end{align}

\subsection{Closed-Form Solution}

\paragraph{Step 1: Compute Centroids}

Compute centroids of the point sets:
\begin{align}
    \bar{\bx} &= \frac{1}{N} \sum_{i=1}^{N} \bx_i, \\[5pt]
    \bar{\by} &= \frac{1}{N} \sum_{i=1}^{N} \by_i.
\end{align}

\paragraph{Step 2: Compute Covariance Matrix}

Define the covariance matrix:
\begin{align}
    \bH = \sum_{i=1}^{N} (\bx_i - \bar{\bx})(\by_i - \bar{\by})^T.
\end{align}

\paragraph{Step 3: Compute Optimal Rotation \( \bR \) via Kabsch-Umeyama}

Perform Singular Value Decomposition (SVD) on \(\bH\):
\begin{align}
    \bH = \bU \mathbf{S} \bV^T.
\end{align}

The optimal rotation from Umeyama~\cite{umeyama1991least} is:
\begin{align}
    \bR = \bV\,\mathrm{diag}(1, 1, \det(\bV \bU^T))\,\bU^T,
\end{align}
ensuring \( \bR \) is a proper rotation (\( \det(\bR) = +1 \)).

\paragraph{Step 4: Compute Symmetric Scale \( s \) via Horn's Method}

Horn's symmetric scale~\cite{horn1988closed} is given by:
\begin{align}
    s = \frac{\sum_{i=1}^{N} \|\by_i - \bar{\by}\|^2}{\sum_{i=1}^{N} \|\bx_i - \bar{\bx}\|^2}.
\end{align}

This symmetric scale factor ensures good performance in forward and inverse alignment tasks.

\paragraph{Step 5: Compute Translation \( \bt \)}

Finally, compute the translation vector:
\begin{align}
    \bt = \bar{\by} - s \bR \bar{\bx}.
\end{align}

\vspace{5pt}

This hybrid approach, combining the rotation and translation from the Kabsch-Umeyama method~\cite{umeyama1991least}
with Horn’s symmetric scale estimation~\cite{horn1988closed}, provides improved accuracy and robustness, using the feature-guided correspondences.

\end{document}